\documentclass[conference]{IEEEtran}
\IEEEoverridecommandlockouts
\usepackage{cite}
\usepackage{amsmath,amssymb,amsfonts}
\usepackage{algorithmic}
\usepackage{graphicx}
\usepackage{textcomp}
\usepackage{xcolor}
\usepackage{array}
\usepackage{booktabs}
\usepackage{multirow}
\usepackage{geometry}
\usepackage{amsthm}

\newtheorem{definition}{Definition}
\newtheorem{lemma}{Lemma}

\def\BibTeX{{\rm B\kern-.05em{\sc i\kern-.025em b}\kern-.08em
    T\kern-.1667em\lower.7ex\hbox{E}\kern-.125emX}}
    
\usepackage{enumitem}
\usepackage{tikz}
\usetikzlibrary{shapes.geometric, arrows.meta, positioning, calc, backgrounds, fit, shadows.blur, patterns}
\usepackage{adjustbox}
\usepackage[T1]{fontenc}
\usepackage{subcaption}

\begin{document}

\definecolor{genColor}{HTML}{009688} 
\definecolor{discColor}{HTML}{FF9800} 
\definecolor{specColor}{HTML}{3F51B5} 
\definecolor{dataColor}{HTML}{607D8B} 

\title{Sequential RC-TGAN: Generating Relational Time Series with Spectral Envelope Loss \\
\thanks{This work was supported by Mitacs through the Mitacs Accelerate program.}
}


\author{
\IEEEauthorblockN{Mohamed Gueye\textsuperscript{1,2}, Yazid Attabi\textsuperscript{1}, Manuel Morales\textsuperscript{2}, and Maxime Dumas\textsuperscript{1}}\\
\IEEEauthorblockA{\textsuperscript{1}\textit{Croesus Lab, Croesus}, Laval, Québec, Canada \\
\textsuperscript{2}\textit{Dept. of Mathematics and Statistics, University of Montreal}, Montréal, Canada \\
\{mohamed.gueye, yazid.attabi, maxime.dumas\}@croesus.com, \{mohamed.gueye, manuel.morales\}@umontreal.ca}
}

\maketitle

\begin{abstract}
The generation of synthetic relational databases often involves modeling complex temporal dynamics, such as transaction logs or event sequences. A significant challenge in this domain is the handling of categorical time series (e.g., status codes), where standard encoding methods like one-hot encoding fail to capture intrinsic frequency-domain features such as seasonality and cyclicity. In this paper, we introduce Sequential RC-TGAN (Seq. RC-TGAN), a temporal extension of the RC-TGAN framework, equipped with a novel integrated loss function based on the \textit{Spectral Envelope Theory}. This differentiable loss allows the generator to directly optimize the preservation of latent periodic structures via backpropagation. While spectral envelope theory is inherently designed for categorical sequences, we extend this frequency-domain regularization to continuous time series by employing a Variational Gaussian Mixture Model (VGM) discretization strategy. To establish a mathematically rigorous evaluation standard, we simulate categorical time series governed by a parameter $\alpha$, with exactly known theoretical spectral envelopes. Integrating these dynamic sequences into the child tables of a relational database yields a robust ground-truth benchmark for evaluating the frequency-domain fidelity of our generative framework. Furthermore, we address the lack of robust evaluation standards for relational time series by proposing two new metrics: Spectral Density Divergence and Spectral Envelope Divergence. Experimental results on real-world datasets, as well as our simulated benchmarks, demonstrate that our end-to-end approach significantly outperforms state-of-the-art systems in reproducing cyclic patterns and long-term seasonality across both categorical and continuous features.

\end{abstract}


\section{Introduction}
\IEEEPARstart{S}ynthetic data generation has rapidly evolved from a niche privacy-preserving technique into a foundational pillar of modern machine learning, addressing critical bottlenecks related to data scarcity and algorithmic fairness while circumventing stringent privacy regulations. Early generative paradigms primarily focused on static, single-table tabular data \cite{xu_modeling_2019, kotelnikov_tabddpm_2024, shi_comprehensive_2025}. Architectures such as TabGPT \cite{padhi2021tabular} and Tabular Transformer GAN (TT-GAN) \cite{kang2025tabular} adapted NLP techniques to generate tabular rows via autoregressive next-token prediction.

In practice, contemporary enterprise data is predominantly structured within complex Relational Databases (RDBs), consisting of interconnected networks of tables governed by strict primary key (PK) and foreign key (FK) constraints. Consequently, multi-table generation models have emerged to address this structural complexity. Early approaches to relational data generation relied on statistical baselines, such as the Synthetic Data Vault (SDV) \cite{patki_synthetic_2016}, which utilized hierarchical Gaussian copulas to model cross-table distributions. The transition to deep learning in this domain was pioneered by the Row Conditional Tabular GAN (RC-TGAN) \cite{gueye_row_2023}, which leveraged Generative Adversarial Networks (GANs) to explicitly maintain referential integrity between parent and child tables. More recently, diverse architectures have been introduced, including transformer-based sequence-to-sequence models like REaLTabFormer \cite{solatorio_realtabformer_2023}, standard tabular diffusion models like ClavaDDPM \cite{pang_clavaddpm_2024}, and graph-based diffusion frameworks such as RelDiff \cite{hudovernik2025reldiff}. Despite their structural sophistication, these relational models are fundamentally static. They treat data as fixed snapshots and obliterate the complex longitudinal dynamics of multivariate time series (e.g., financial transaction logs) embedded within these schemas.

Synthesizing dynamic temporal sequences that are structurally embedded within a relational database requires conditioning child time series trajectories on the static parent table. Models such as TimeGAN \cite{yoon2019time} and DoppelGANger \cite{lin2020using} pioneered this domain. However, these methods operate strictly in the time domain and are not focused on categorical time series, which are ubiquitous in real-world relational tables. Because standard representations like one-hot encoding map categories to orthogonal, equidistant vectors (where the Euclidean distance is always $\sqrt{2}$), the neural network becomes completely blind to ordinal, hierarchical, or periodic relationships, preventing the generator from understanding the cyclical nature of categorical states. Apprehending these complex cyclical patterns in the time domain is inherently difficult, often leading to models missing crucial structural amplitudes. Consequently, enriching this sequence analysis with the frequency domain provides a significantly better approach, allowing the model to explicitly uncover and optimize the latent periodic structures underlying the discrete categories.

To overcome the intersecting limitations of time-domain optimization and categorical data modeling, we propose a profound paradigm shift by directly integrating \textit{Spectral Envelope Theory} \cite{stoffer1993spectral, stoffer2000spectral, shumway2006time} into a relational generative architecture. We introduce Sequential RC-TGAN (Seq. RC-TGAN), equipped with a novel, differentiable spectral envelope loss that explicitly exploits the frequency domain to optimize the generator's ability to model the complex pattern of categorical time series. The principle of spectral envelope in this context is to find an optimal scalar transformation that maximizes the spectral density of categorical time series; through this principle, we successfully translate discrete categories into continuous numerical representations. While recent literature has also pivoted toward frequency-domain regularization with architectures such as the Frequency-Markov Diffusion GAN (FMD-GAN) \cite{ma2026dynamic}, FDEDiff \cite{zhangfrequency}, and TIFO \cite{piao2026tifo} introducing highly innovative frequency-aware denoising, these methods are primarily designed for single-table generation and continuous time series, processing categorical
time series using standard one-hot encoding representations. Consequently, they fail to resolve the challenge of categorical periodicity, as they cannot natively assign spectral meaning to orthogonal vectors without extensive feature engineering.

Our main contributions are as follows:
\begin{itemize} 
	\item We integrate spectral envelope theory into a conditional sequential GAN framework by introducing a novel spectral loss term ($\mathcal{L}_{spec}$). This loss explicitly minimizes the distance between the spectral envelopes of real and synthetic data, overcoming the orthogonality of one-hot encodings to preserve latent periodic structures in categorical time series. 
	\item We extend this spectral methodology to continuous numerical features by employing beforehand a discretization strategy based on Gaussian Mixture Models (GMM) \cite{xu_modeling_2019}, allowing the spectral envelope to capture and enforce frequency-domain features across mixed data types simultaneously.
	\item We analytically derive the exact theoretical spectral envelope for Markov chains governed by circulant transition matrices. This provides a mathematically tractable and rigorous "gold standard" benchmark simulated dataset to evaluate the frequency-domain fidelity of sequential generative models without relying on empirical periodograms.
	\item We propose a new set of evaluation metrics rooted in spectral analysis : Spectral Density Divergence ($\overline{\mathcal{D}}_{spec}$) and Spectral Envelope Divergence ($\overline{\mathcal{D}}_{env}$). These metrics are designed to rigorously assess the temporal fidelity and cyclic consistency of generated  continuous and categorical time series, addressing the blindspots of traditional time-domain metrics.
\end{itemize}

The remainder of this paper is organized as follows. Section \ref{sec:background_spec_analysis} provides the necessary background on spectral analysis. Section \ref{sec:metric_space_formulation} formulates spectral envelopes within a metric space. Section \ref{sec:sequential_rctgan} introduces the proposed GAN framework for multi-table time series synthesis. Section \ref{sec:markov_chains} outlines the design of simulated data for our experiments. Section \ref{sec:new_metrics_specenv} defines the new evaluation metrics based on spectral analysis. Finally, Section \ref{sec:experiments} presents the experimental setup and results, followed by concluding remarks.

\section{Background on Spectral Analysis} \label{sec:background_spec_analysis}
\subsection{Spectral Density}
Let $\{S_t, t \in \mathbb{Z}\}$ be a weakly stationary process with value on $\mathbb{R}$, with mean $\mu \in \mathbb{R}$ and autocovariance function $\gamma(h) = \text{cov}(S_{t+h}, S_t)$. The spectral density $f(\omega)$ describes how the variance of the process is distributed across frequencies $\omega \in [-1/2, 1/2]$. The spectral density is defined as the Fourier transform of the autocovariance:

\begin{equation}\label{equ:spectral_dens_decomposition}
    f(\omega) = \sum_{h=-\infty}^{\infty} \gamma(h) e^{-2\pi i \omega h}.
\end{equation}
Inversely : $\gamma(h) = \int_{-1/2}^{1/2}{f(\omega) e^{2\pi i \omega h} d\omega}.$\\

In practice, for a finite time series $\{s_1, \dots, s_T\}$, the spectral density is estimated using the periodogram $I(\omega_k)$, calculated at Fourier frequencies $\omega_k = k/T$:
$$
I(\omega_k) = \left| d(\omega_k) \right|^2 = \frac{1}{T} \left| \sum_{t=1}^{T} s_t e^{-2\pi i \omega_k t} \right|^2,
$$
where $d(\omega_k) = \frac{1}{\sqrt{T}} \sum_{t=1}^{\prime} s_t e^{-2\pi i \omega_k t}$, the Discrete Fourier Transform (DFT). 

The concept of spectral density can be extended to the multivariate case. Let $\{\textbf{S}_t\}_{t \in \mathbb{Z}}$ be a weakly stationary process in $\mathbb{R}^q$ with mean $\mu$ and autocovariance matrix $\Gamma(h) = \mathbb{E}[(\textbf{S}_{t+h} - \mu)(\textbf{S}_t - \mu)^{\prime}]$. The $(j, p)$-th entry of this matrix is the \textit{cross-covariance function} $\gamma_{jp}(h) = \text{cov}(\textbf{S}_{j, t+h}, \textbf{S}_{p, t})$, which measures the covariance between component $j$ at time $t+h$ and component $p$ at time $t$.\\
The spectral density matrix $\textbf{f}(\omega) \in \mathbb{C}^{q \times q}$ is defined as the Fourier transform of $\Gamma(h)$:
$$
    \textbf{f}(\omega) = \sum_{h=-\infty}^{\infty} \Gamma(h) e^{-2\pi i \omega h}, \quad \omega \in [-1/2, 1/2].
$$
The diagonal elements $\textbf{f}_{jj}(\omega)$ represent the univariate spectral densities, while the off-diagonal elements $\textbf{f}_{jp}(\omega)$ denote the cross-spectral densities. \\
For a finite observation $\{\textbf{s}_1, \dots, \textbf{s}_T\}$, the spectral density is estimated using the periodogram. Let $\textbf{d}(\omega_k) \in \mathbb{C}^q$ be the Discrete Fourier Transform (DFT) at frequency $\omega_k = k/T$:
$$
    \textbf{d}(\omega_k) = \frac{1}{\sqrt{T}} \sum_{t=1}^{T} \textbf{s}_t e^{-2\pi i \omega_k t}.
$$
The multivariate periodogram matrix $\textbf{I}(\omega_k)$ is defined as the outer product:
\begin{equation}
    \textbf{I}(\omega_k) = \textbf{d}(\omega_k) \textbf{d}(\omega_k)^*,
\end{equation}
where $^*$ denotes the conjugate transpose. While $\textbf{I}(\omega_k)$ is an asymptotically unbiased estimator of $\textbf{f}(\omega_k)$, it is not consistent; its variance does not vanish as $T \to \infty$. Consequently, consistent estimation requires smoothing techniques, such as windowing or averaging over frequency bands.

\subsection{Spectral Envelope for Categorical Time Series}
Consider a categorical time series $X_t$ taking values in a finite set $\textbf{a} = \{a_0, \dots, a_{K-1}\}$ that is stationary. Because standard frequency-domain tools cannot be directly applied to discrete qualitative symbols, we assign a vector of numerical scaling values $\beta = (\beta_0, \dots, \beta_{K-1})^{\prime} \in \mathbb{R}^K$ to the categories in $\textbf{a}$. This transformation results in a real-valued numerical process, denoted $X_t(\beta) \in \mathbb{R}$, where $X_t(\beta) = \beta_k$ whenever the original series is in state $X_t = a_k$. By explicitly mapping the qualitative categories to quantitative scalars, we convert the discrete sequence into a standard univariate continuous-state time series. This mathematical conversion is a strict prerequisite, as it enables the calculation of autocovariance functions and the subsequent computation of the spectral density via the Fourier transform.

Instead of assigning arbitrary numbers to categories, the spectral envelope framework systematically derives optimal numerical values that expose hidden periodicities within a categorical time series. The primary objective is to find a scaling vector $\beta$ that maximizes the spectral density relative to the total variance at each specific frequency $\omega$. Formally, the \textit{Spectral Envelope} $\lambda(\omega)$ is defined as:
\begin{equation}\label{equ:spectral_envelope_def}
    \lambda(\omega) = \sup_{\beta \not\propto \mathbf{1}} \left\{ \frac{f(\omega; \beta)}{\sigma^2(\beta)} \right\}, \quad \forall \omega \in [-1/2, 1/2],
\end{equation}
where $f(\omega; \beta)$ and $\sigma^2(\beta)$ represent the spectral density and the variance of the transformed numerical process $X_t(\beta)$, respectively \cite{stoffer1993spectral}. The condition $\beta \not\propto \mathbf{1}$ explicitly excludes trivial scalings where every category is assigned the exact same numerical value. If $\beta$ is proportional to a vector of all ones ($\beta \propto \mathbf{1}$), the transformed sequence $X_t(\beta)$ would merely become a flat, constant series. This would result in a variance of zero ($\sigma^2(\beta) = 0$), thereby rendering the objective ratio undefined.

This optimization problem can be solved by representing the categorical process as a multivariate point process $Y_t \in \mathbb{R}^K$ (using one-hot vectors). Let $f_Y(\omega)$ be the spectral density matrix and $V_Y$ be the variance matrix of stationary process $Y_t$. The optimization problem in (\ref{equ:spectral_envelope_def}) can be re-written :
\begin{equation}
    \lambda(\omega) = \sup_{\beta \not\propto \mathbf{1}} \left\{ \frac{\beta^{\prime}f_Y(\omega)\beta}{\beta^{\prime}V_Y\beta} \right\}, \quad \forall \omega \in [-1/2, 1/2].
\end{equation}
This expression is a generalized Rayleigh quotient. The solution $\lambda(\omega)$ is the largest eigenvalue of $f_Y(\omega)$ in the metric of $V_Y$. The corresponding eigenvector $\beta(\omega)$ is called the optimal scaling at frequency $\omega$.\\
The value $\lambda(\omega)$ is called the \textit{spectral envelope} because it envelopes the normalized spectrum of any scaled process $X_t(\beta)$. In other words, for any normalized scaling $\beta$ (such that $\sigma^2(\beta)=1$), we have $f(\omega; \beta) \le \lambda(\omega)$, with equality achieved if and only if $\beta$ is proportional to the optimal scaling $\beta(\omega)$.


While the spectral envelope provides a robust mechanism for uncovering the latent periodicities of a single categorical time series, leveraging this concept within a deep generative framework requires systematically comparing the structural properties of real and synthesized processes. To formulate a differentiable objective that minimizes the frequency-domain discrepancy between these temporal dynamics, we cannot merely view $\lambda(\omega)$ as a collection of point-wise maxima. Instead, we must formalize the spectral envelope as a distinct mathematical object residing within a well-defined functional space. This theoretical shift naturally motivates the construction of a metric space for spectral envelopes, providing the foundational distance metrics required to optimize our generative model via backpropagation.

\section{A Metric Space Formulation for Spectral Envelopes} \label{sec:metric_space_formulation}
Consider a stationary categorical process $X_t$ taking values in the finite set $\textbf{a}$ with spectral envelope $\lambda(\omega)$. Let $X^{(\theta)}_t$ be a parametric stationary categorical process (e.g., a synthetic process generated by a model) with values in $\textbf{a}$, parameters $\theta$, and spectral envelope $\lambda^{(\theta)}(\omega)$. 

In a generative context, the goal is to ensure that the synthetic process $X^{(\theta)}_t$ approximates the real process $X_t$. A fundamental question arises: how can we quantify the discrepancy between these processes in the frequency domain? By defining a metric distance between $\lambda(\omega)$ and $\lambda^{(\theta)}(\omega)$, we can formulate an optimization problem where minimizing this distance with respect to $\theta$ forces the synthetic process to recover the latent periodic structures of the real data. We first formalize the space in which these spectral envelopes reside.

\begin{definition}
    Let $\mathcal{S}_K$ be the set of spectral envelopes corresponding to stationary categorical processes with $K$ categories that possess a continuous spectral density matrix associated with their one-hot encoding representation (i.e. $f_Y(\omega)$).
\end{definition}

\begin{lemma}\label{lemma:specenv_continuity} 
    Every element $\lambda \in \mathcal{S}_K$ is a continuous function on the interval $[-1/2, 1/2]$. \\
    \textit{The proof of this lemma is in the appendix.}
\end{lemma}

Consequently, $\mathcal{S}_K$ is a subset of $C^0\left([-1/2, 1/2]\right)$, the space of continuous functions on the fundamental frequency domain. This inclusion implies that $\mathcal{S}_K$ resides within the Hilbert space $L^2\left([-1/2, 1/2]\right)$.

The Hilbert space $L^2\left([-1/2, 1/2]\right)$ consists of square-integrable functions defined on $[-1/2, 1/2]$ equipped with the inner product:
$$
    \langle h, g \rangle = \int_{-1/2}^{1/2}{h(\omega) g(\omega) d\omega}, \quad \forall h, g \in L^2.
$$
This induces the $L^2$ norm, representing the total energy of the function:
$$
    \Vert h \Vert_2 = \sqrt{\langle h, h \rangle}  = \sqrt{\int_{-1/2}^{1/2}{h(\omega)^2 d\omega}}.
$$

From this functional space definition, we derive metrics to measure the distance between the real spectral envelope $\lambda$ and the synthetic spectral envelope $\lambda^{(\theta)}$:
\begin{equation}\label{equ:l_2_dist_specenv}
    \Vert \lambda - \lambda^{(\theta)} \Vert_2 = \sqrt{\int_{-1/2}^{1/2}{\left(\lambda(\omega) - \lambda^{(\theta)}(\omega)\right)^2 d\omega}}
\end{equation}

The $L^2$ distance in (\ref{equ:l_2_dist_specenv}) aggregates the error over the entire frequency domain. It is differentiable (assuming $\lambda^{(\theta)}$ is differentiable with respect to $\theta$) and provides non-zero gradients for deviations across all frequencies simultaneously. This "smoothness" makes the $L^2$ metric significantly tractable as a loss function for backpropagation in deep neural networks. Therefore, we adopt the square of the $L^2$ distance as our objective function to minimize the divergence between the real and synthetic spectral envelopes.

\begin{lemma}\label{lemma:specenv_boundness}
    For all $\lambda \in \mathcal{S}_K$, the following norm properties hold:
    \begin{enumerate}[label=(\roman*), nosep]
        \item $1 \leq \Vert \lambda \Vert_1 \leq K-1$.
        \item $1 \leq \Vert \lambda \Vert_2 < \infty$.
    \end{enumerate}
    \textit{The proof of this lemma is in the appendix.}
\end{lemma}

The $L^1$ upper bound ($\Vert \lambda \Vert_1 \leq K-1$) reflects the dimensionality constraint of a categorical variable with $K$ states, where the rank of the associated variance-covariance matrix is at most $K-1$. Regarding the lower bound $\Vert \lambda \Vert_2 \geq 1$, this property constitutes a fundamental energy constraint for any non-trivial stationary process. Because the spectral density $f(\omega)$ decomposes the total variance of the process across frequencies, the integral of $f(\omega)$ must equal the variance $\gamma(0)$. Given that $\lambda(\omega)$ is defined as the supremum that envelopes the normalized spectrum of any scaled process, its integral (the $L^1$ norm) cannot be less than the variance of a standardized process ($\sigma^2 = 1$). By the relationship between norms on a compact domain of length 1, we have $\Vert \lambda \Vert_2 \geq \Vert \lambda \Vert_1 \geq 1$. This lower bound represents the "white noise" baseline where the spectral mass is uniformly distributed. In the context of deep learning, this ensures the loss function is anchored; the generator cannot minimize the spectral distance by simply reducing the synthetic process to a trivial or zero-variance state, as it must maintain the minimum spectral energy inherent to a categorical distribution.

\section{Conditional GAN for Multi-table Time Series Synthesis}
\label{sec:sequential_rctgan}
\subsection{Formalization and Notation}
Our formulation is grounded in the Probabilistic Relational Model (PRM) framework \cite{koller1999probabilistic}. We consider a relational schema $\mathcal{S} = \{ W, U \}$ containing two classes (tables) $W$ and $U$, where $W$ acts as the parent entity and $U$ as the child entity. 

Let $\mathcal{A}(U)$ denote the set of attributes for table $U$. This set is partitioned into continuous attributes $\mathcal{A}_{cont}(U) = \{c_1, \dots, c_I\}$ and categorical attributes $\mathcal{A}_{cat}(U) = \{d_1, \dots, d_J\}$. The attribute space (or domain) for $U$ is defined as the Cartesian product of the domains of its individual attributes: $\mathcal{V}(U) = \bigotimes_{A \in \mathcal{A}(U)} \mathcal{V}(A)$. Similarly, we define $\mathcal{V}(W)$ as the attribute space for the parent table $W$.

In this relational structure, specific dependencies exist between instances of $W$ and $U$. Let $w \in W$ denote a specific row (instance) in the parent table, with feature values $w.\mathcal{A} \in \mathcal{V}(W)$. We define $\text{Children}(w) \subset U$ as the set of child rows in table $U$ that reference the parent $w$. 

In the context of time series synthesis, the set $\text{Children}(w)$ is not merely a bag of rows but an ordered sequence associated with the parent entity. We denote this sequence as $\text{Children}(w).\mathcal{A} = (u_1, \dots, u_T)$, where each $u_t \in \mathcal{V}(U)$ represents the state of the child entity at time step $t$, and $T$ is the sequence length. Thus, the dataset consists of tuples $ \left(w.\mathcal{A}, \text{Children}(w).\mathcal{A} \right)$, pairing static parent features with dynamic child sequences.

\subsection{Sequential RC-TGAN Architecture}
\begin{figure}
    \centering
    \includegraphics[width=1.0\linewidth]{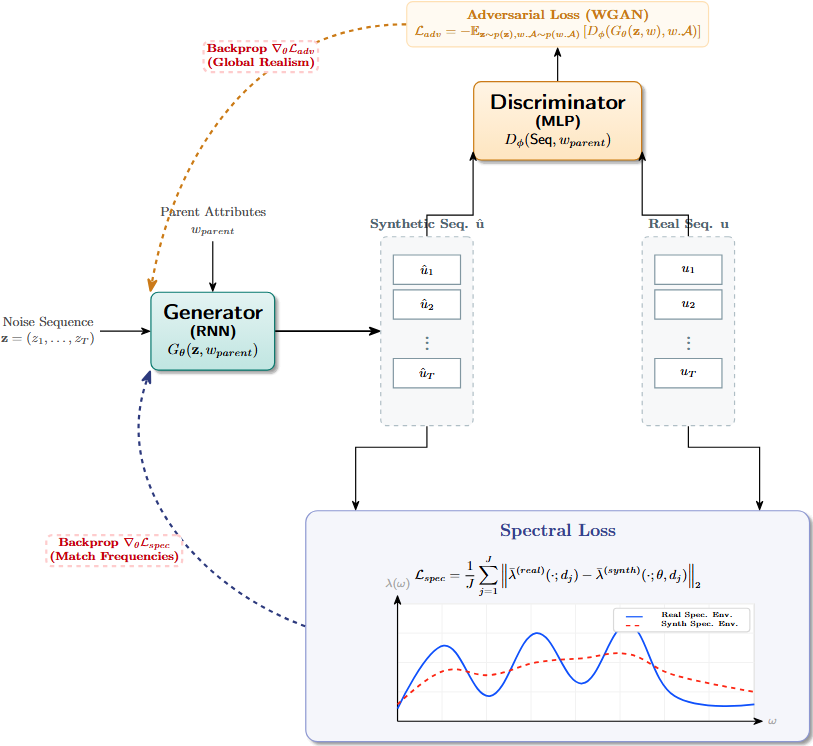}
    \caption{Architecture schema of the Sequential RC-TGAN with Spectral Loss. The diagram illustrates the generation process conditioned on parent attributes, and the dual optimization setup where the generator receives adversarial feedback from the discriminator and frequency-domain feedback via the spectral envelope loss ($\mathcal{L}_{spec}$).}
    \label{fig:seq_rctgan_schema}
\end{figure}

To address the challenge of generating relational time series, we introduce the \textbf{Sequential RC-TGAN} (see Fig. \ref{fig:seq_rctgan_schema}), an extension of the Row Conditional-TGAN (RC-TGAN) \cite{gueye_row_2023} model enhanced by the temporal dimension modeling. The original RC-TGAN primarily focused on modeling inter-table relationships, employing a generator $\mathcal{G}$ to model the conditional distribution of a single child row given its parent: $\mathbb{P}(u | w.\mathcal{A})$.

The Sequential RC-TGAN adapts this paradigm to support inter-row relationships modeling inside a tabular data. Rather than mapping parent feature values $w.\mathcal{A}$ and a noise vector $z$ to a static point in the feature space $\mathcal{V}(U)$, our generator learns to map them to a temporal trajectory within $\mathcal{V}(U)^{T}$. Formally, the model approximates the conditional joint distribution of the child sequence given the parent attributes:
\begin{equation}
    \mathbb{P}(\text{Children}(w).\mathcal{A} | w.\mathcal{A}) = \mathbb{P}(u_1, \dots, u_T | w.\mathcal{A}).
\end{equation}
This formulation ensures that the generation process is explicitly conditioned on the static characteristics of the parent entity, thereby guaranteeing that the synthesized temporal dynamics remain consistent with their relational context.

\subsubsection{Conditional Recurrent Generator}
To capture temporal dependencies effectively, we replace the fully connected layers of the original RC-TGAN with a Recurrent Neural Network (RNN) generator \cite{medsker2001recurrent}.

The generation process is conditioned on the static parent attributes $w.\mathcal{A}$ at every time step, ensuring the generated sequence adheres to the specific constraints of the parent entity. At each time step $t$, the generator receives a concatenated input consisting of a random noise vector $z_t \sim \mathcal{N}(0, I)$ and the parent vector $w.\mathcal{A}$:
\begin{align}
    h_t &= \text{RNN}(h_{t-1}, [z_t \oplus w.\mathcal{A}]) \\
    \hat{u}_t &= \text{MLP}(h_t)
\end{align}
where $\oplus$ denotes concatenation, $h_t$ represents the hidden state, and $\hat{u}_t$ is the generated attribute vector at time $t$. By reinjecting $w.\mathcal{A}$ at each step, this architecture ensures that the static relational constraints (e.g., Store Type, Location) exert a persistent influence over the entire dynamic trajectory of the child sequence.

\subsubsection{Conditional Discriminator}
In contrast to the generator, the discriminator $D_\phi$ is implemented as a fully connected network (MLP) designed to assess the global coherence of the sequence. It models the joint probability of the entire sequence conditioned on the parent attributes.

Assuming a fixed sequence length $T$ during training, the input to the discriminator is constructed by flattening the sequence $\{u_1, \dots, u_T\}$ into a single vector and concatenating it with the parent attributes $w.\mathcal{A}$. The discriminator then maps this joint representation $[u_1 \oplus \dots \oplus u_T \oplus w.\mathcal{A}]$ to a validity score, determining whether the complete temporal trajectory constitutes a plausible instance given the specific parent context.

\subsection{Spectral Adaptation for Continuous Features} \label{subsec:spectral_adaptation_for_continuous_feat}
The spectral envelope theory in \cite{stoffer1993spectral} is inherently designed for categorical time series. However, relational datasets frequently contain continuous numerical attributes $\mathcal{A}_{cont}(U) = \{c_1, \dots, c_I\}$ that exhibit significant periodic behavior (e.g., sales volume, temperature). To incorporate these attributes into our frequency-domain regularization, we first employ a discretization strategy based on Variational Gaussian Mixture Models (VGM) \cite{xu_modeling_2019}.

For each continuous attribute $c_i \in \mathcal{A}_{cont}(U)$, we fit a VGM to the training data to estimate the optimal number of modes $K_{c_i}$ and their parameters. The probability distribution of a value $u_{t, c_i}$ is modeled as a mixture of Gaussians:
\begin{equation}
    \mathbb{P}(u_{t, c_i}) = \sum_{k=1}^{K_{c_i}} \tau_k \mathcal{N}(u_{t, c_i}; \mu_k, \sigma_k).
\end{equation}

To compute the spectral envelope for a continuous sequence $\mathbf{u}_{c_i} = (u_{1, c_i}, \dots, u_{T, c_i})$, we transform it into a discrete sequence of mode indicators $\mathbf{m}_{c_i} = (m_{1, c_i}, \dots, m_{T, c_i})$. At each time step $t$, the value $u_{t, c_i}$ is assigned to the mode $k$ that maximizes the posterior probability:
\begin{equation}
    m_{t, c_i} = \arg\max_{k} \left( \tau_k \mathcal{N}(u_{t, c_i}; \mu_k, \sigma_k) \right).
\end{equation}

This process effectively maps the continuous domain $\mathcal{V}(c_i)$ to a finite categorical set $\{1, \dots, K_{c_i}\}$. Consequently, we can calculate the spectral envelope $\lambda(\omega; c_i)$ on this discretized sequence, allowing the spectral loss $\mathcal{L}_{spec}$ to enforce periodic consistency across both naturally categorical attributes $\mathcal{A}_{cat}(U)$ and discretized continuous attributes $\mathcal{A}_{cont}(U)$.

Beyond this discrete mode assignment, each continuous value is concurrently represented by a normalized scalar that captures its relative position within the assigned mode. Specifically, if the value $u_{t, c_i}$ is assigned to mode $k$, we compute an intra-mode scalar $v_{t, c_i} = \frac{u_{t, c_i} - \mu_k}{4\sigma_k}$. By concatenating the one-hot encoded discrete mode indicator $m_{t, c_i}$ with this normalized continuous scalar $v_{t, c_i}$, the model retains the complete information necessary to fully reconstruct the original continuous feature $u_{t, c_i}$. Therefore, while the categorical mode sequence $\mathbf{m}_{c_i}$ explicitly drives the frequency-domain regularization via the spectral envelope, the supplementary scalar sequence $v_{t, c_i}$ ensures no loss of localized continuous variance in the time domain.

Note that another way to incorporate the continuous attributes into frequency domain is to use the power spectrum of the signal. However as it will be shown in the ablation study section, we find that  the discretization method is more effective.

\subsection{Generator Losses}
The training of the generator is guided by a hybrid objective function designed to satisfy two complementary requirements: global statistical realism (via adversarial feedback) and frequency-domain fidelity (via spectral envelope matching).

\subsubsection{Adversarial Loss ($\mathcal{L}_{adv}$)}
The primary objective of the generator is to produce relational sequences that are indistinguishable from real data. To achieve stable training dynamics, we employ the Wasserstein GAN (WGAN) objective.

Let $\mathbb{P}_r$ denote the real data distribution and $\mathbb{P}_g$ the generator distribution conditioned on parent attributes $w.\mathcal{A}$. The discriminator $D$ (or critic) aims to maximize the divergence between its scoring of real and synthetic sequences. Conversely, the generator $G$ minimizes this divergence. The adversarial loss for the generator is defined as:
\begin{equation}
    \mathcal{L}_{adv} = -\mathbb{E}_{\textbf{z} \sim p(\textbf{z}), w.\mathcal{A} \sim p(w.\mathcal{A})} \left[D_\phi(G_{\theta}(\textbf{z}, w), w.\mathcal{A})\right].
\end{equation}
Minimizing this term encourages the generator to capture general temporal correlations and the joint distribution of the sequence conditioned on the parent $w$.

\subsubsection{Spectral Envelope Loss ($\mathcal{L}_{spec}$)}
Standard adversarial losses often fail to capture frequency patterns in categorical time series because discriminators tend to focus on local transitions rather than global frequency structures. To remedy this, we introduce a regularization term based on the spectral envelope.

\textbf{Sequence-wise Spectral Estimation:}
Since the spectral envelope is a statistical property, we estimate it over mini-batches to ensure stability. Let $\mathcal{B} = \{ \mathbf{u}^{(1)}, \dots, \mathbf{u}^{(B)} \}$ be a mini-batch of $B$ sequences. For a specific categorical feature $d_j$, we compute the spectral envelope $\lambda\left(\omega; \mathbf{u}^{(b)}_{d_j}\right)$ for the $b$-th sequence at frequency $\omega$ (as defined in (\ref{equ:spectral_envelope_def})).

We calculate the \textit{mean spectral envelope} for the real batch, $\bar{\lambda}^{(real)}$, and the synthetic batch, $\bar{\lambda}^{(synth)}$, by averaging the envelopes across the batch dimension:
\begin{equation}\label{equ:mean_specenv_cat}
    \bar{\lambda}^{(real)}(\omega; d_j) = \frac{1}{B} \sum_{b=1}^{B} \lambda\left(\omega; \mathbf{u}^{(b)}_{d_j}\right).
\end{equation}
This batch-averaging step reduces the variance of the periodogram estimator and provides a robust target frequency profile for the generator. We adapt this estimation for the continuous numerical features $c_i$ by relying on their discrete mode indicators (as detailed in Section \ref{subsec:spectral_adaptation_for_continuous_feat}). Let $\mathbf{m}^{(b)}_{c_i}$ denote the discretized sequence for the $b$-th instance of feature $c_i$. The mean spectral envelope is correspondingly calculated as:
\begin{equation}\label{equ:mean_specenv_cont}
    \bar{\lambda}^{(real)}(\omega; c_i) = \frac{1}{B} \sum_{b=1}^{B} \lambda\left(\omega; \mathbf{m}^{(b)}_{c_i}\right).
\end{equation}
The synthetic counterparts, $\bar{\lambda}^{(synth)}(\omega; d_j)$ and $\bar{\lambda}^{(synth)}(\omega; c_i)$, are computed analogously over the generated batch.

\begin{figure}
    \centering
    \includegraphics[width=1.0\linewidth]{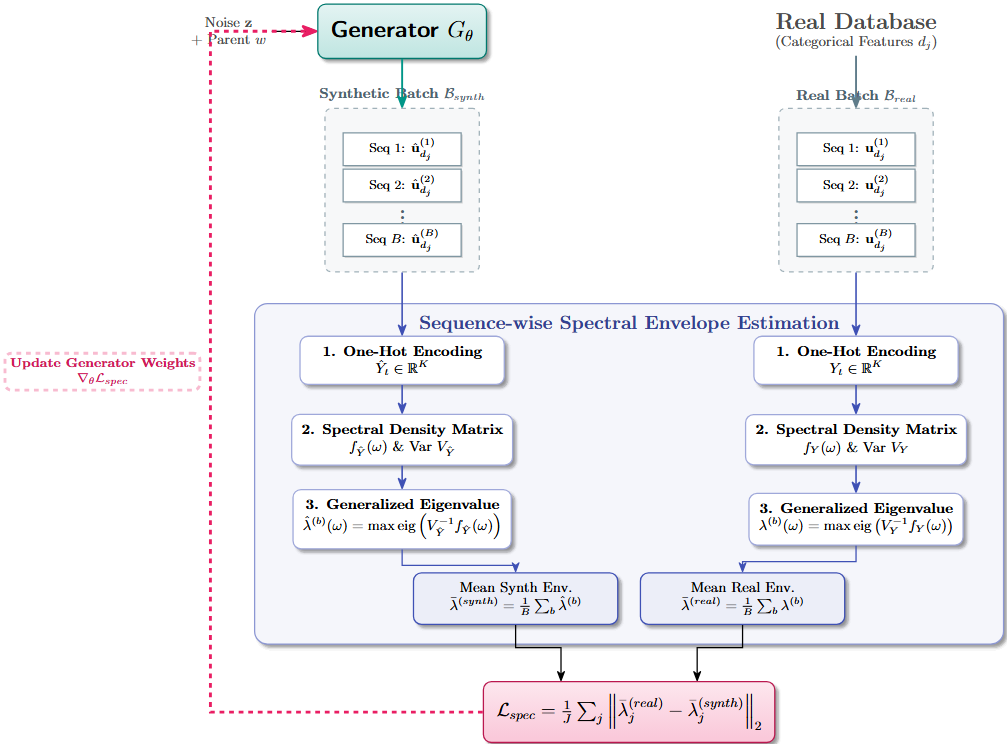}
    \caption{Detailed training flow of the generator via the Spectral Envelope Loss. Categorical sequences from both real and synthetic mini-batches undergo one-hot encoding to estimate their respective multivariate spectral density and variance matrices. The spectral envelopes are derived by solving the generalized eigenvalue problem. The loss explicitly minimizes the $L_2$ distance between the batch-averaged envelopes, providing continuous, differentiable frequency-domain gradients ($\nabla_\theta \mathcal{L}_{spec}$) to update the recurrent generator.}
    \label{fig:generator_training_spec_loss}
\end{figure}

\textbf{Loss Formulation:}
To enforce periodic consistency across the entire relational dataset, we partition our frequency-domain objective into two components. The categorical spectral loss, $\mathcal{L}_{spec}^{(cat)}$, minimizes the average $L_2$ distance between the real and synthetic mean spectral envelopes across all $J$ categorical features:
\begin{equation}\label{equ:l_spec_cat}
    \mathcal{L}_{spec}^{(cat)} = \frac{1}{J}\sum_{j=1}^{J} \left\Vert \bar{\lambda}^{(real)}(\cdot; d_j) -  \bar{\lambda}^{(synth)}(\cdot; \theta, d_j) \right\Vert_2.
\end{equation}
Likewise, the continuous spectral loss, $\mathcal{L}_{spec}^{(cont)}$, computes the average $L_2$ distance across the $I$ discretized numerical features:
\begin{equation}\label{equ:l_spec_cont}
    \mathcal{L}_{spec}^{(cont)} = \frac{1}{I}\sum_{i=1}^{I} \left\Vert \bar{\lambda}^{(real)}(\cdot; c_i) -  \bar{\lambda}^{(synth)}(\cdot; \theta, c_i) \right\Vert_2.
\end{equation}
The total spectral envelope loss, $\mathcal{L}_{spec}$, is constructed as the weighted sum of these two terms, distributed proportionally to the number of features of each type:
\begin{equation}\label{equ:l_spec_total}
    \mathcal{L}_{spec} = \frac{J}{J+I} \mathcal{L}_{spec}^{(cat)} + \frac{I}{J+I} \mathcal{L}_{spec}^{(cont)}.
\end{equation}
Minimizing this unified term explicitly forces the generator to align the latent periodicities (e.g., seasonality, cyclic trends) of the synthetic sequences with the ground truth across both mixed data types (see Fig. \ref{fig:generator_training_spec_loss}).

\subsection{Training Loop}

The training procedure employs an alternating optimization strategy to balance the competing objectives. In each epoch, we execute the following three distinct phases:

\begin{enumerate}
    \item \textbf{Discriminator Update:} First, we optimize the discriminator $D$ to distinguish between real sequences and the current synthetic output. We perform $n_{critic}$ updates to the discriminator for every generator update to maintain an optimal gradient approximation for the WGAN objective.

    \item \textbf{Adversarial Generator Update:} Second, we update the generator $G$ by minimizing $\mathcal{L}_{adv}$. In this step, the generator weights are adjusted to fool the discriminator, ensuring global statistical coherence and adherence to the parent conditioning.

    \item \textbf{Spectral Generator Update:} Finally, we perform a specialized refinement step focused on frequency-domain fidelity. We update the generator by minimizing $\mathcal{L}_{spec}$. This update is repeated $n_{steps\_for\_spec}$ times per epoch.
\end{enumerate}


\section{Design of Simulated Data for Experiments}
\label{sec:markov_chains}

Validating generative models on real-world categorical time series is inherently difficult. Because real-world data lacks a definitive "ground truth" for its underlying stochastic frequencies, evaluations often rely on noisy periodogram estimates. To rigorously evaluate whether a generative model genuinely learns complex frequency-domain features, rather than merely memorizing local transitions, it is crucial to employ benchmark time series where the spectral properties are known beforehand.

To this end, we turn to stationary Markov chains. These stochastic processes provide a controlled, "gold standard" evaluation environment for two primary reasons: first, they can be easily and exactly simulated to generate massive, customized datasets for model training; second, they allow for the exact analytical derivation of their theoretical spectral envelope. By comparing the empirical spectral envelope of the generated sequences against this mathematically known ground truth, we can accurately measure the frequency fidelity of our synthetic approximations.

We map the categorical series $X_t$ into the multivariate point process $Y_t$ (one-hot vector). Specifically, $Y_t$ takes values in the set of standard basis vectors $\{e_0, \dots, e_{K-1}\} \subset \mathbb{R}^K$. In this one-hot encoded representation, $e_j$ is a vector with a $1$ at the $j+1$-th position and $0$ everywhere else, corresponding exactly to the event that $X_t$ is in state $a_j$. The process is characterized by the \textit{Transition Matrix Function}, denoted as $\mathcal{T}(h)$. This matrix-valued function describes the conditional probability of the process transitioning from one basis state to another over a given time lag $h \ge 1$.

For a stationary categorical process, the entry $(i, j)$ of the transition matrix function at lag $h$, $\mathcal{T}_{ij}(h)$, represents the probability of transitioning from state $e_i$ to state $e_j$ after $h$ steps:
\begin{equation}
    \mathcal{T}_{ij}(h) = \mathbb{P}[Y_{t+h} = e_j \mid Y_t = e_i].
\end{equation}
This matrix captures the "flow" of probability mass across the state space over time.

In the context of a first-order Markov chain, the behavior of the transition matrix function $\mathcal{T}(h)$ is strictly governed by the immediate 1-step transitions. Consequently, the transition probabilities at any lag $h$ are entirely determined by the $h$-th power of the 1-step transition matrix $P = \mathcal{T}(1)$:
$$
    \mathcal{T}(h) = P^h.
$$

This transition relationship is the key to computing the temporal covariance of the process. For a stationary categorical process $Y_t$ characterized by a stationary distribution vector $\pi$ (row vector), let $\Pi = \text{diag}(\pi)$ denote the diagonal matrix of its marginal probabilities. The autocovariance matrix function, $\Gamma(h)$, is directly related to the transition matrix function $\mathcal{T}(h)$ by the following equation:
\begin{equation} \label{equ:markov_autocov}
    \Gamma(h) = \Pi \mathcal{T}(h) - \pi^{\prime} \pi, \quad \text{for } h \ge 0.
\end{equation}

By substituting the property established above for a first-order Markov chain where the multi-step transition is simply the matrix power $\mathcal{T}(h) = P^h$, this general relation simplifies significantly. The autocovariance matrix function reduces to a geometric decay governed entirely by the 1-step transition matrix $P$:
$$
    \Gamma(h) = \Pi P^h - \pi^{\prime} \pi.
$$

Equation (\ref{equ:markov_autocov}) reveals that the spectral properties of the process are entirely governed by the relaxation of the transition mechanism $\mathcal{T}(h)$.

For the remainder of this section, we assume that the stationary categorical process $X_t$ is a first-order Markov chain characterized by the one-step transition matrix $P = \mathcal{T}(1)$.

\subsection{Spectral Properties of Circulant Transitions}
Deriving the spectral envelope for a general transition matrix requires numerically solving the eigenvalue problem at every frequency. For the class of \textit{circulant} transition matrices, we can derive an exact analytical form that links the stochastic parameters directly to the spectral shape.

A transition matrix $P$ is \textit{circulant} if every row is a cyclic right shift of the preceding row. Consequently, the entire matrix is fully characterized by its first row vector $\textbf{b} = [b_0, b_1, \dots, b_{K-1}]$, where $b_j = \mathbb{P}(Y_{t+1}=e_j \mid Y_t=e_0)$.

The general form of such a matrix is:
\begin{equation}\label{equ:circulant_matrix}
    P = \begin{bmatrix}
        b_0     & b_1     & b_2    & \dots  & b_{K-1} \\
        b_{K-1} & b_0     & b_1    & \dots  & b_{K-2} \\
        b_{K-2} & b_{K-1} & b_0    & \dots  & b_{K-3} \\
        \vdots  & \vdots  & \vdots & \ddots & \vdots  \\
        b_1     & b_2     & b_3    & \dots  & b_0
    \end{bmatrix}.
\end{equation}
This structural symmetry serves as a mathematical bridge between the time domain and the frequency domain: circulant matrices are diagonalized by the Inverse Discrete Fourier Transform matrix \cite{kra2012circulant}, a property we leverage to derive analytical spectral envelopes. Then, the eigenvalues $\gamma_k$ of the circulant matrix in (\ref{equ:circulant_matrix}) is given by:
\begin{equation}
    \gamma_k = \sum_{j=0}^{K-1} b_j e^{i \frac{2\pi j k}{K}}\text{ for } k=0,\ldots, K-1.
\end{equation}

The magnitude ($|\gamma_k|$) is determined by the concentration of the probability mass in $\textbf{b}$. If $\textbf{b}$ is highly concentrated (low entropy), the magnitude approaches 1 ($|\gamma_k| \approx 1$), implying long memory, whereas a uniform $\textbf{b}$ (high entropy) yields $|\gamma_k| \approx 0$, which is characteristic of a white noise process.

\begin{lemma}[Spectral Envelope of Circulant Chains]
\label{lemma:circulant_envelope}
Let $X_t$ be a stationary categorical process with $K$ states governed by a circulant transition matrix $P$. Let $\gamma_k = r_k e^{i\phi_k}$ be the eigenvalues of $P$ expressed in polar form i.e. $r_k = |\gamma_k|$ and $\phi_k = \arg(\gamma_k)$. The spectral envelope $\lambda(\omega)$ is the upper boundary of the spectral densities of the $K-1$ non-trivial eigenmodes:

\begin{equation}\label{equ:circulant_envelope}
    \lambda(\omega) = \max_{k \in \{1, \dots, K-1\}} \left( \frac{1 - r_k^2}{1 - 2r_k \cos(2\pi\omega - \phi_k) + r_k^2} \right).
\end{equation}
\textit{The proof of this lemma is in the appendix.}
\end{lemma}

Using this lemma, we analyze two types of circulant chains representing distinct temporal dynamics: periodicity and inertia.

\subsection{The Noisy Cyclic Process (Periodicity)}
This process models periodic behavior with phase noise, serving as a robust benchmark for capturing seasonality and cyclic constraints.

A \textit{Noisy Cyclic Process} (NCP) is defined by the transition matrix:
$$ P_{ij} = \begin{cases} \alpha & \text{if } j \equiv (i+1) \pmod K \\ 1-\alpha & \text{if } j=i \end{cases} $$
where $\alpha \in (0.5, 1)$ is the \textit{switching state} parameter. An NCP is a circulant chain where $\textbf{b} = [1-\alpha, \alpha, 0, \dots]$. Its spectral envelope is given by equation (\ref{equ:circulant_envelope}) where : 
\begin{align*}
    r_k &= \sqrt{1 - 2\alpha(1-\alpha)\left[1 - \cos\left(\frac{2\pi k}{K}\right)\right]}, \\
    \phi_k &= \arctan\left( \frac{\alpha \sin(2\pi k / K)}{(1-\alpha) + \alpha \cos(2\pi k / K)} \right).
\end{align*}
The NCP serves as a robust benchmark for modeling periodic behavior and cyclic constraints under varying degrees of phase noise. Its temporal dynamics are primarily controlled by the switching state parameter $\alpha \in (0.5, 1)$, which dictates the strictness of the cycle. Figure \ref{fig:spectral_envelope_circulant_matrix} visualizes the spectral envelope of the NCP with a state space of $K=7$ for different values of $\alpha$. As shown, the process naturally exhibits distinct resonant peaks clustered around the fundamental frequency of $1/7 \approx 0.14$ and its associated harmonics. When the cycle strength $\alpha$ approaches 1 (represented by the darker lines), the system mimics a deterministic cycle, concentrating the spectral energy into very sharp, Dirac-like peaks. Conversely, as $\alpha$ decreases toward its lower bound, the process introduces greater phase noise, which progressively broadens these sharp harmonic peaks into wide spectral hills, reflecting a more stochastic and relaxed periodic progression.

\subsection{The Symmetric Sticky Process (Inertia)}
This process models systems with inertia, where the state tends to persist over time with no preferred direction of change.

A \textit{Symmetric Sticky Process} (SSP) is defined by the transition matrix:
$$ P_{ij} = \begin{cases} \alpha & \text{if } i=j \\ \frac{1-\alpha}{K-1} & \text{if } i \neq j \end{cases} $$
where $\alpha \in (1/K, 1)$ is the \textit{switching state} parameter. We can remark that the transition matrix of the SSP is circulant such that the first row $\textbf{b} = [\alpha, \frac{1-\alpha}{K-1}, \dots]$. We can derive the spectral envelope of the SSP from equation (\ref{equ:circulant_envelope}):
$$
    \lambda(\omega) = \frac{1 - \gamma^2}{1 - 2\gamma \cos(2\pi\omega) + \gamma^2},
$$
where $\gamma = \frac{\alpha K - 1}{K - 1}$. The non-trivial eigenvalues are identical and real corresponding to $\gamma$.

\begin{figure}
    \centering
    \includegraphics[width=0.8\linewidth]{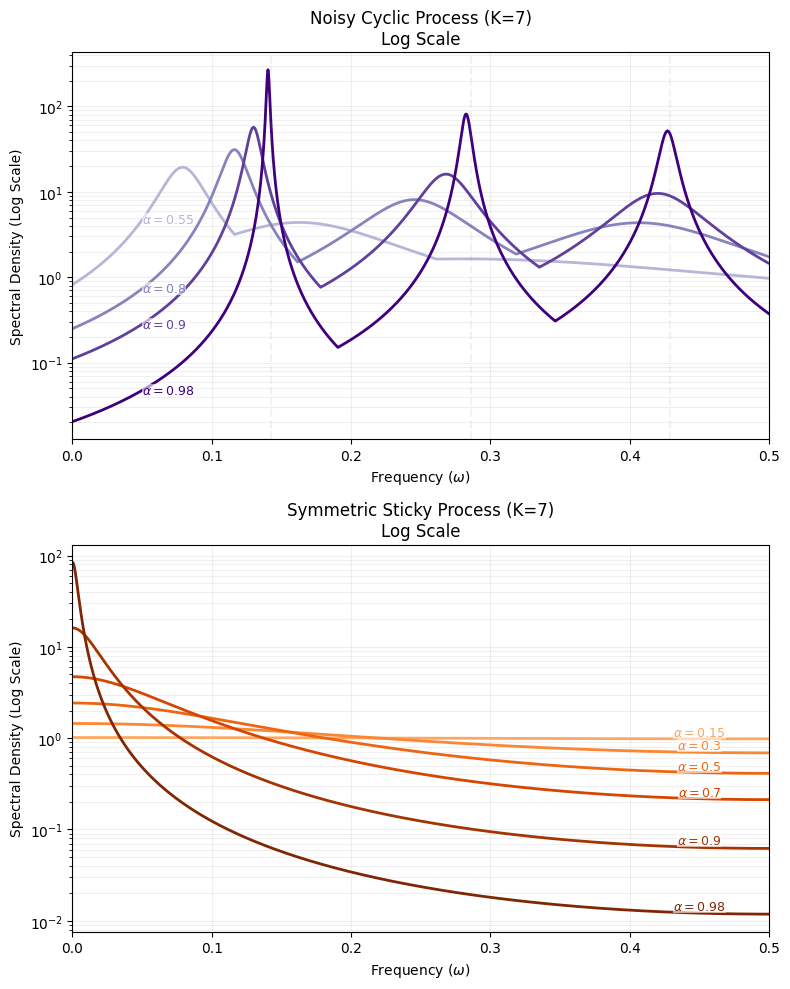}
    \caption{\textbf{Spectral envelopes of the benchmark Markov chains ($K=7$) on a logarithmic scale.} 
    \textbf{Upper:} The NCP exhibits peaks close to the fundamental frequency ($1/7 \approx 0.14$) and its harmonics. As the cycle strength $\alpha \to 1$ (darker lines), the process approaches a deterministic clock with Dirac-like peaks. Lower $\alpha$ values introduce phase noise, broadening the peaks into wide spectral hills. The log scale is used to visualize the large dynamic range between stochastic and near-deterministic regimes.
    \textbf{Bottom:} The SSP acts as a low-pass filter. As the persistence $\alpha \to 1$ (darker lines), the spectral energy concentrates strictly at $\omega=0$, representing system inertia. Lower $\alpha$ values result in a flat, white-noise-like spectrum. 
    }
    \label{fig:spectral_envelope_circulant_matrix}
\end{figure}

The asymptotic behavior of the process is highly sensitive to the switching state parameter $\alpha$. As $\alpha \to 1$, the eigenvalue $\gamma \to 1$, causing the process to become extremely "sticky" and rarely switch states. In this regime, the spectral envelope forms a sharp peak at $\omega=0$, ultimately approaching a Dirac delta. Conversely, as $\alpha \to 1/K$, the eigenvalue $\gamma \to 0$, which reduces the process to pure random noise (see Figure \ref{fig:spectral_envelope_circulant_matrix}). Consequently, the spectral energy becomes uniformly distributed, and the envelope flattens to a constant line where $\lambda(\omega) \approx 1$.

\subsection{Bayesian Hierarchical Sampling for generating simulated relational databases}
To rigorously evaluate the conditional generation capabilities of our model, we elaborate a method for building a synthetic relational database using a Bayesian hierarchical framework. Unlike  using a single fixed parameter for the entire dataset, we model the switching parameter $\alpha$ as a random variable associated with each parent entity. This setup forces the generative model to learn the mapping $\alpha \mapsto \text{Spectral Envelope}(\alpha)$ rather than memorizing a static distribution.



Following the relational schema $\mathcal{S}$ introduced in Section \ref{sec:sequential_rctgan}, we construct a framework where the parent table $W$ governs the stochastic dynamics of the child time series in table $U$. Note that for any single simulated database, we select only one of these two types of circulant chains (either the Symmetric Sticky or the NCP) and fix the total number of categorical states $K$ to drive the temporal dynamics. Because both processes are fully parameterized by the switching parameter $\alpha$ for a given $K$, we apply a Bayesian hierarchical modeling to each choice: first, we sample and store $N$ values of $\alpha$ in the parent table; second, we generate a corresponding categorical time series for each of these $N$ values using the chosen Markov process.

For the NCP, where $\alpha$ represents the probability of advancing the cycle, we use a uniform prior over the valid range of directed cycles:
$$
    \alpha_i \sim \mathcal{U}(0.5, 1.0),
$$
where $\alpha_i$ is the random variable illustrating the value of $\alpha$ for the $i$-th row in the parent table. For the Symmetric Sticky, we simply replace $\mathcal{U}(0.5, 1.0)$ distribution by $\mathcal{U}(1/K, 1.0)$.

For each parent row $i$, we generate a categorical time series $\textbf{u}^{(i)} = \{u^{(i)}_1, \dots, u^{(i)}_T\}$ of length $T$, which acts as part of the child table $X$. The dynamics of this series are conditioned strictly on the parent's parameter $\alpha_i$.

The sequence is generated via the transition matrix $P_{\alpha_i}$ specific to the chosen type of circulant chain (Symmetric Sticky or Noisy Cyclic):
\begin{equation}
    \mathbb{P}(u^{(i)}_{t+1} | u^{(i)}_t, \alpha_i) = [P_{\alpha_i}]_{u^{(i)}_t, u^{(i)}_{t+1}}.
\end{equation}

This hierarchical construction provides a "gold standard" dataset for conditional generative modeling. Since the true spectral envelope $\lambda(\omega; \alpha_i)$ is analytically known for every parent $i$ (via Lemma \ref{lemma:circulant_envelope}), we can compute the exact expected spectral error. A successful generative model must produce synthetic children $\hat{\textbf{u}}^{(i)}$ such that their empirical spectral envelopes match the theoretical envelopes dictated by their sampled parent attributes $\hat{\alpha}_i$.

\section{New Metrics Based on Spectral Analysis}
\label{sec:new_metrics_specenv}

Evaluating the fidelity of synthetic relational time series requires going beyond simple marginal distributions or static correlations. Standard metrics often fail to accurately detect if the synthetic data preserves the specific frequency-domain characteristics (such as seasonality and cyclic constraints) inherent to the real process. To address this, we propose two metrics rooted in spectral analysis.

Let $\{w_1, \dots, w_M\}$ be the set of parent instances in the real database, and $\{\hat{w}_1, \dots, \hat{w}_{M'}\}$ be the set of parent instances in the synthetic database. Each parent $w_m$ (or $\hat{w}_{m'}$) identifies a specific sub-population of children rows forming a multivariate time series.

For a numerical attribute $c_i$, let $f(\omega; c_i, w_m)$ denote the spectral density of the series associated with parent $w_m$. For a categorical attribute $d_j$, let $\lambda(\omega; d_j, w_m)$ denote its spectral envelope. We define the mean spectral densities and mean spectral envelopes for the real data as:

\begin{align}
    \bar{f}^{(real)}(\omega; c_i) & = \frac{1}{M}\sum_{m=1}^{M} f(\omega; c_i, w_m),\\
    \bar{\lambda}^{(real)}(\omega; d_j) & = \frac{1}{M}\sum_{m=1}^{M} \lambda(\omega; d_j, w_m).
\end{align}

The synthetic counterparts, $\bar{f}^{(synth)}$ and $\bar{\lambda}^{(synth)}$, are defined analogously over the $M'$ synthetic parents.

\subsection{Spectral Density Divergence ($\overline{\mathcal{D}}_{spec}$)}
To evaluate the temporal fidelity of continuous features, we measure the divergence between the average power spectrums. We first normalize the spectral densities so that $\int_{-1/2}^{1/2} f(\omega) d\omega = 1$, considering them as probability distributions over the frequency domain.

The individual divergence $div(\cdot, \cdot)$ for an attribute $c_i$ is defined as the divergence (e.g., Wasserstein or Kullback-Leibler) between the real and synthetic mean densities. The global \textit{Spectral Density Divergence} (SDD) is the average over all continuous attributes:
\begin{equation}
    \label{equ:spec_dens_metrics}
    \overline{\mathcal{D}}_{spec} = \frac{1}{I} \sum_{i=1}^{I}{\mathcal{D}_{spec}(c_i)},
\end{equation}
where $\mathcal{D}_{spec}(c_i) = div \left( \bar{f}^{(real)}(\cdot; c_i), \bar{f}^{(synth)}(\cdot; c_i) \right)$. It is worth noting that the aggregated quantities $\bar{f}^{(real)}(\cdot; c_i)$ and $\bar{f}^{(synth)}(\cdot; c_i)$ constitute well-defined normalized spectral densities. Mathematically, the set of valid normalized spectral densities is closed under convex combinations; since each individual $f(\omega; c_i, w_m)$ is a non-negative, real-valued, its integral under interval $[-1/2, 1/2]$ is equal to one, and even function (for real-valued processes), their arithmetic mean preserves these fundamental properties. Consequently, $\bar{f}^{(real)}$ effectively represents the spectral density of a "representative" process for the class $U$, averaging out local idiosyncrasies to reveal the global frequency structure of the population. This validity allows us to treat these patterns as probability distributions and rigorously apply divergence metrics $div$ such as Kullback-Leibler (KL) or Wasserstein distance.

\subsection{Spectral Envelope Divergence ($\overline{\mathcal{D}}_{env}$)}
For categorical attributes, we assess the preservation of latent periodicities using the spectral envelope. The divergence for a single attribute $d_j$ is calculated as the $L^2$ distance between the mean envelopes:
\begin{equation}
    \mathcal{D}_{env}(d_j) = \frac{\big\Vert \bar{\lambda}^{(real)}(\cdot; d_j) - \bar{\lambda}^{(synth)}(\cdot; d_j) \big\Vert_2}{K-1}.
\end{equation}
The global \textit{Spectral Envelope Divergence} (SED) is the mean over all categorical attributes:
\begin{equation}
    \label{equ:specenv_metrics}
    \overline{\mathcal{D}}_{env} = \frac{1}{J} \sum_{j=1}^{J} \mathcal{D}_{env}(d_j).
\end{equation}

The aggregated spectral envelope $\bar{\lambda}^{(real)}(\omega; d_j)$ inherits the fundamental algebraic properties of its constituents. Since $\bar{\lambda}^{(real)}$ is constructed as a finite linear combination of individual envelopes, and given that each component $\lambda(\cdot; d_j, w_m)$ is continuous on the compact interval $[-1/2, 1/2]$ (Lemma \ref{lemma:specenv_continuity}), the mean envelope itself is necessarily a continuous function in $C^0([-1/2, 1/2])$. Furthermore, the fundamental norm constraints established in Lemma \ref{lemma:specenv_boundness} are preserved under this averaging operation. Specifically, by the convexity of the norm, the mean envelope satisfies the dimensionality constraint $1 \leq \Vert \bar{\lambda}^{(real)}(\cdot; d_j) \Vert_1 \leq K-1$ (If $K$ is the number of categories of $d_j$) and maintains finite energy $1 \leq \Vert \bar{\lambda}^{(real)}(\cdot; d_j) \Vert_2 < \infty$. These properties ensure that $\bar{\lambda}^{(real)}$ remains a well-defined term within the functional $L^2\left([-1/2, 1/2]\right)$ and verify important properties of spectral envelopes, guaranteeing that the divergence metric $\overline{\mathcal{D}}_{env}$ is both bounded and mathematically stable.


\section{Experiments} \label{sec:experiments}
In this section, we empirically evaluate the performance of our proposed generative framework. The primary objective is to demonstrate that integrating the spectral envelope loss into a sequential GAN architecture significantly improves the preservation of latent periodicities and temporal dynamics in both continuous and categorical time series conveyed by rows of child tables of relational databases. To this end, we conduct comprehensive experiments on both simulated data and real-world transactional datasets. We compare our model against several state-of-the-art generative approaches, followed by an ablation study to evaluate the contribution of specific components within our proposed model.

\subsection{Experimental setup}

\subsubsection{Simulated Data (Bayesian Hierarchical Benchmarks)}
We first evaluate our models on simulated relational databases generated via our Bayesian Hierarchical Benchmarking framework. We simulate two distinct categorical Markov processes:
\begin{itemize}
    \item NCP: Tests the model's ability to capture periodic behaviors of categorical time series. The transition parameter is sampled from the prior $\alpha \sim \mathcal{U}(0.5, 1.0)$.
    \item SSP: Tests the model's ability to capture low-frequency, high-inertia dynamics (low-pass filter behavior). The transition parameter is sampled from the prior $\alpha \sim \mathcal{U}(1/K, 1.0)$.
\end{itemize}
Specifically, we synthesize a parent table comprising 100 independent entities (rows), where each entity's attribute $\alpha$ is drawn from the corresponding uniform prior. For each parent row, we simulate a sequence of 10,000 child rows, generated using the Markov transition matrix governed by that specific $\alpha$. This hierarchical construction dictates the exact theoretical spectral envelope of the generated child sequences, providing a rigorous "gold standard" to isolate and evaluate frequency-domain fidelity across varying state space sizes ($K=7, 12, 21$).

\subsubsection{Real-World Datasets}
We also evaluate our method on two real-world relational databases containing complex temporal dynamics: Rossmann \cite{noauthor_rossmann_nodate} and Walmart \cite{tommy_wilczek_walmart_nodate}. Each database adheres to a two-level hierarchical relational schema. The parent table contains static metadata representing the set of individual stores (e.g., store type, location), while the child table contains the multivariate time series for each store, encompassing both continuous numerical features (e.g., daily or weekly sales volume) and highly periodic categorical features (e.g., day of the week, promotional events). 

\subsubsection{Baseline models}
We compare our proposed method (Seq. RC-TGAN) against state-of-the-art systems spanning standard relational generation and dedicated time series generative adversarial networks:
\begin{itemize}
    \item SDV \cite{patki_synthetic_2016}: A standard probabilistic relational model that builds generative models of relational databases by computing statistics at the intersection of related tables.
    \item ClavaDDPM \cite{pang_clavaddpm_2024}: A recent diffusion-based approach for multi-relational data synthesis.
    \item DoppelGANger \cite{lin2020using}: A state-of-the-art GAN designed for networked time series that tackles mode collapse and long-term dependencies.
    \item TimeGAN \cite{yoon2019time}: A time series GAN that combines the unsupervised adversarial paradigm with the control of supervised training through a jointly optimized latent embedding space.
\end{itemize}

\subsubsection{Metrics}
To rigorously assess both temporal fidelity and the preservation of complex temporal patterns conveyed by child tables rows structures, we employ the following metrics:
\begin{itemize}
    \item MSE (ACF): The Mean Squared Error between the autocorrelation of the real and synthetic data. This measures the model's ability to preserve time-domain dependencies and localized temporal structures.
    \item SDD ($\overline{\mathcal{D}}_{spec}$): Defined in (\ref{equ:spec_dens_metrics}), we specifically use the KL divergence as $div$ function to evaluate categorical time series.
    \item SED ($\overline{\mathcal{D}}_{env}$): Defined in (\ref{equ:specenv_metrics}), used  to evaluate numerical time series.  
\end{itemize}

\subsection{Experimental Results}

The results are summarized in Tables \ref{tab:metrics_real_data} and \ref{tab:metrics_simulated_data}, alongside visual analyses in Figures \ref{fig:acf_grid_analysis} and \ref{fig:spectral_simulated_results}. To assess model performance, we highlight relative improvements over the second-best baselines and evaluate statistical significance using a two-sample t-test ($p < 0.05$).

\subsubsection{Performance on Simulated Data} 
Table \ref{tab:metrics_simulated_data} demonstrates that \textit{Seq. RC-TGAN} achieves statistically significant reductions in SED ($\overline{\mathcal{D}}_{env}$) across all tested state space sizes ($K=7, 12, 21$). For the NCP, our proposed method yields relative improvements of 45.8\%, 43.3\%, and 28.2\% over the next best baselines at $K=7, 12,$ and $21$, respectively. This trend holds for the SSP, where \textit{Seq. RC-TGAN} significantly outperforms the second-best models by 32.7\% ($K=7$), 52.8\% ($K=12$) and 37.48\% ($K=21$).

\begin{figure*}[t!]
    \centering
    \begin{subfigure}[b]{0.49\textwidth}
        \centering
        \includegraphics[width=\textwidth]{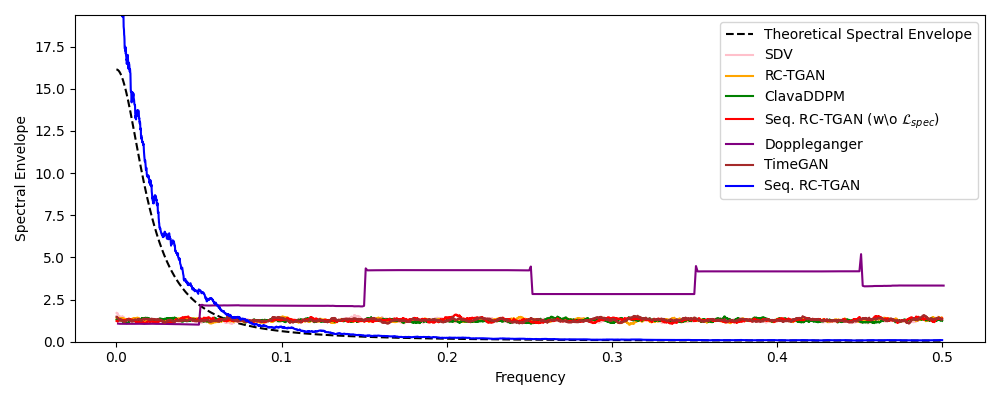}
        \caption{Symmetric Sticky ($\alpha=0.9$)}
        \label{fig:spec_sticky_09}
    \end{subfigure}
    \hfill
    \begin{subfigure}[b]{0.49\textwidth}
        \centering
        \includegraphics[width=\textwidth]{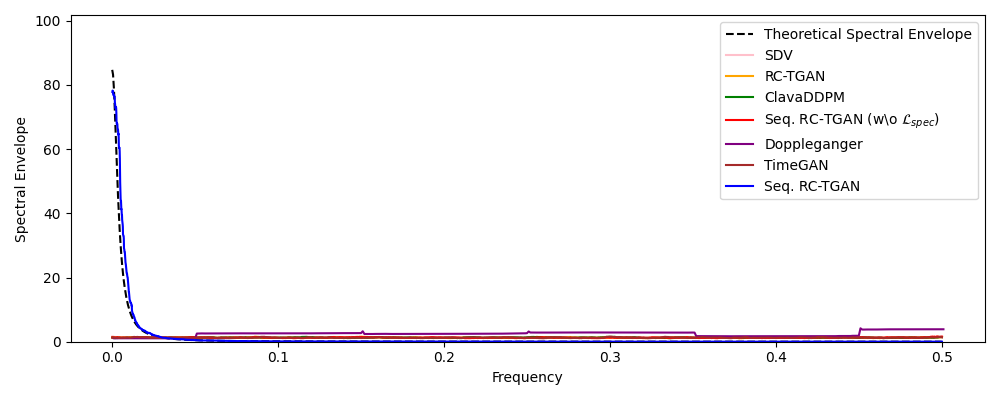}
        \caption{Symmetric Sticky ($\alpha=0.98$)}
        \label{fig:spec_sticky_098}
    \end{subfigure}
    
    \vspace{0.3cm} 
    
    \begin{subfigure}[b]{0.47\textwidth}
        \centering
        \includegraphics[width=\textwidth]{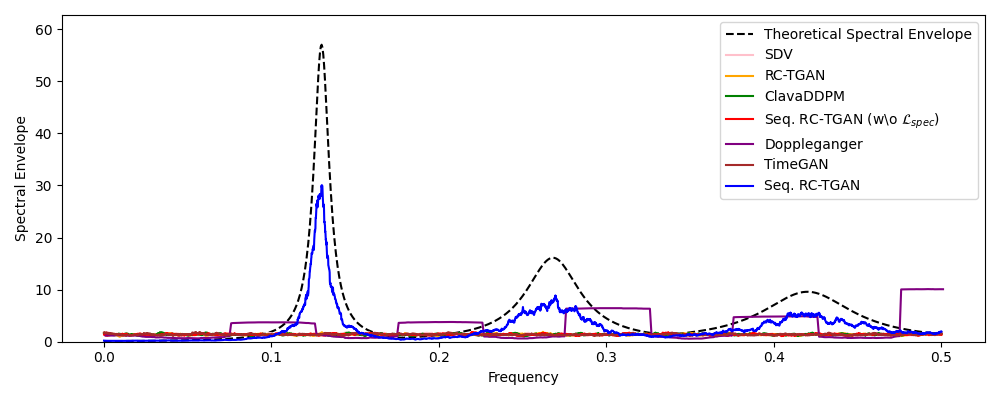}
        \caption{Noisy Cyclic ($\alpha=0.9$)}
        \label{fig:spec_cyclic_09}
    \end{subfigure}
    \hfill
    \begin{subfigure}[b]{0.47\textwidth}
        \centering
        \includegraphics[width=\textwidth]{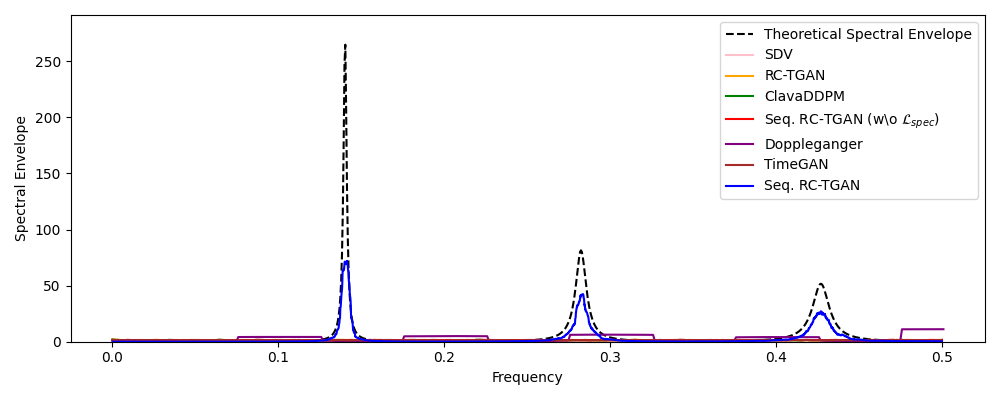}
        \caption{Noisy Cyclic ($\alpha=0.98$)}
        \label{fig:spec_cyclic_098}
    \end{subfigure}

    \caption{\textbf{Spectral Envelope Evaluation on Simulated Data ($K=7$).} The black dashed line represents the theoretical ground truth. \textbf{Top Row (Sticky):} The proposed \textit{Seq. RC-TGAN} (blue) accurately captures the low-pass behavior, with spectral mass concentrating at $\omega=0$ as persistence ($\alpha$) increases. \textbf{Bottom Row (Cyclic):} \textit{Seq. RC-TGAN} successfully aligns with the fundamental harmonic peaks (e.g., $\omega \approx 0.14$) and their sharpening as cycle strength ($\alpha$) increases. Conversely, sequential baselines—including DoppelGANger (purple), TimeGAN (brown), and the unregularized \textit{Seq. RC-TGAN (no $\mathcal{L}_{spec}$)} (red)—fail to capture these periodic structures, collapsing into flat or erratic noise.}
    \label{fig:spectral_simulated_results}
\end{figure*}

Figure \ref{fig:spectral_simulated_results} provides a qualitative spectral analysis of the generated sequences against the mathematical ground truth (black dashed line). A rigorous visual evaluation of synthetic categorical time series in the frequency domain necessitates assessing two critical criteria: (1) \textit{frequency localization}, ensuring the synthetic envelope successfully exhibits peaks at all theoretical fundamental and harmonic frequencies; (2) \textit{spectral purity}, verifying that every peak in the synthetic envelope corresponds to a true theoretical peak without introducing spurious periodic artifacts. 

Our proposed model consistently satisfies both criteria. For the SSP (top row), it accurately captures the theoretical low-pass filter behavior, concentrating the precise magnitude of spectral mass exactly at $\omega=0$ without hallucinating higher-frequency artifacts as the persistence parameter $\alpha$ approaches 1. For the NCP (bottom row), the model successfully isolates the true fundamental harmonic peaks (e.g., near $\omega \approx 0.14$ for $K=7$), maintains strict spectral purity by avoiding false peaks, and faithfully reproduces the theoretical amplitudes to accurately model the phase noise. In stark contrast, competing baselines universally fail these criteria, failing to detect true periodicities and typically collapsing into flat, white-noise-like representations.

\subsubsection{Performance on Real-World Data} 
Table \ref{tab:metrics_real_data} details the performance of the generative models on the Rossmann and Walmart databases. The proposed framework establishes state-of-the-art performance across multiple facets of temporal generation. Most notably, it demonstrates a unique ability to model complex categorical time series, yielding statistically significant divergence reductions in SED ($\overline{\mathcal{D}}_{env}$) compared to the second-best baselines: 37.8\% and 59.3\% relative gain on the Rossmann and Walmart databases, respectively ($p < 0.05$).

Furthermore, although the spectral envelope theory described in section \ref{sec:background_spec_analysis} is inherently designed for categorical sequences, our framework extends this frequency-domain regularization to continuous numerical attributes by employing a Variational Gaussian Mixture Model (VGM) discretization strategy. By mapping continuous values to discrete mode indicators, the spectral loss successfully enforces periodic consistency across all feature types. This adaptation directly contributes to our model achieving a statistically significant 33.7\% relative improvement in continuous SDD ($\overline{\mathcal{D}}_{spec}$) on the Rossmann dataset, as well as a relative gain of 24.8\% over DoppelGANger on the Walmart dataset ($p < 0.05$ for both).

SDV, ClavaDDPM, and TimeGAN consistently demonstrate the weakest performance, struggling to capture both local time-domain structures and global frequency distributions. In contrast, DoppelGANger proves to be a highly competitive state-of-the-art baseline for numerical data. On the Walmart dataset, DoppelGANger and our proposed model are statistically tied for the best time-domain performance (MSE ACF of 0.012 vs. 0.013, respectively; $p > 0.05$). However, despite DoppelGANger's proficiency with continuous variables, it struggles significantly when tasked with modeling the periodic dynamics of categorical time series. Our proposed method overcomes this important limitation via our unified spectral envelope regularization.

\begin{figure*}[t!]
    \centering
    \begin{subfigure}[b]{0.47\textwidth}
        \centering
        \includegraphics[width=\textwidth]{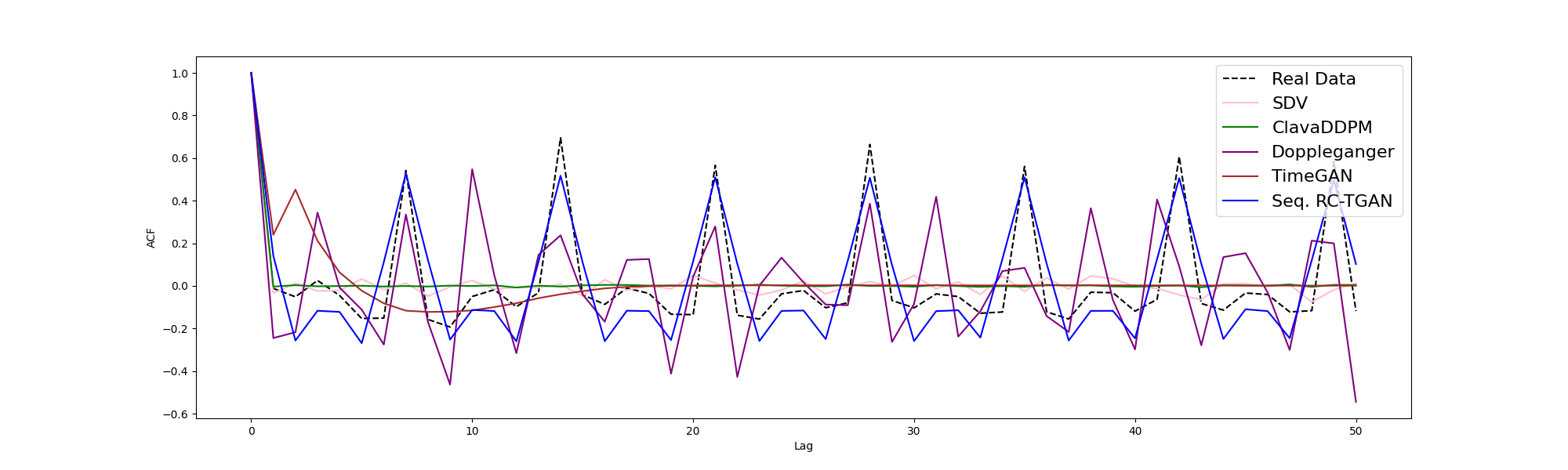}
        \caption{Rossmann Store Sales: Baseline comparison}
        \label{fig:acf_rossman_bench}
    \end{subfigure}
    \hfill
    \begin{subfigure}[b]{0.47\textwidth}
        \centering
        \includegraphics[width=\textwidth]{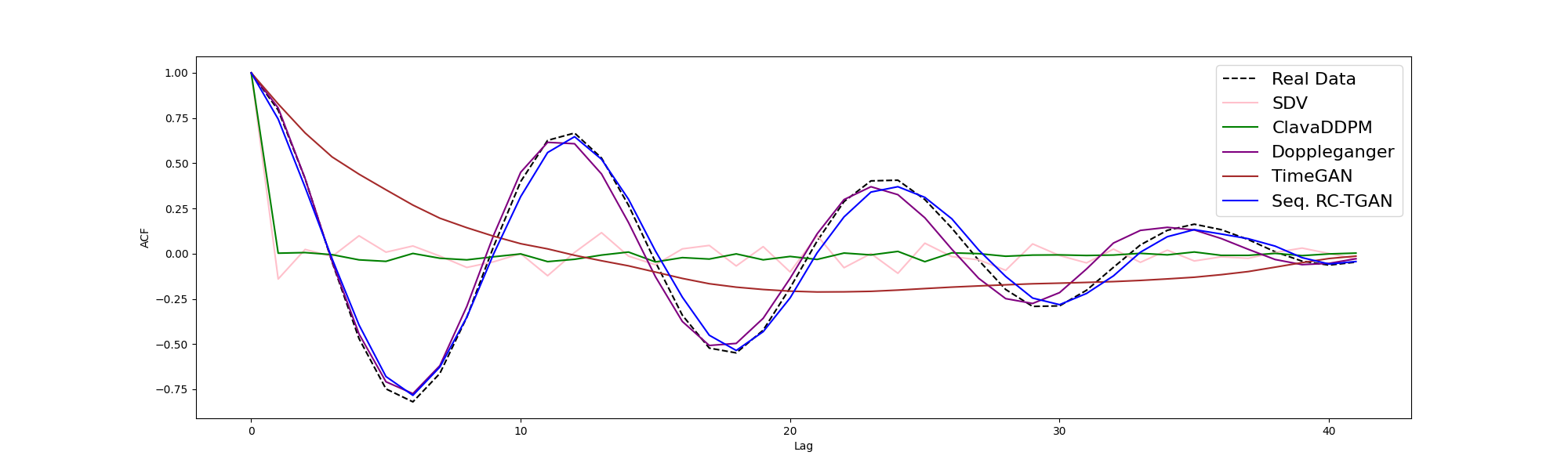}
        \caption{Walmart Fuel Price: Baseline comparison}
        \label{fig:acf_walmart_bench}
    \end{subfigure}
    
    \vspace{0.1cm} 
    
    \begin{subfigure}[b]{0.47\textwidth}
        \centering
        \includegraphics[width=\textwidth]{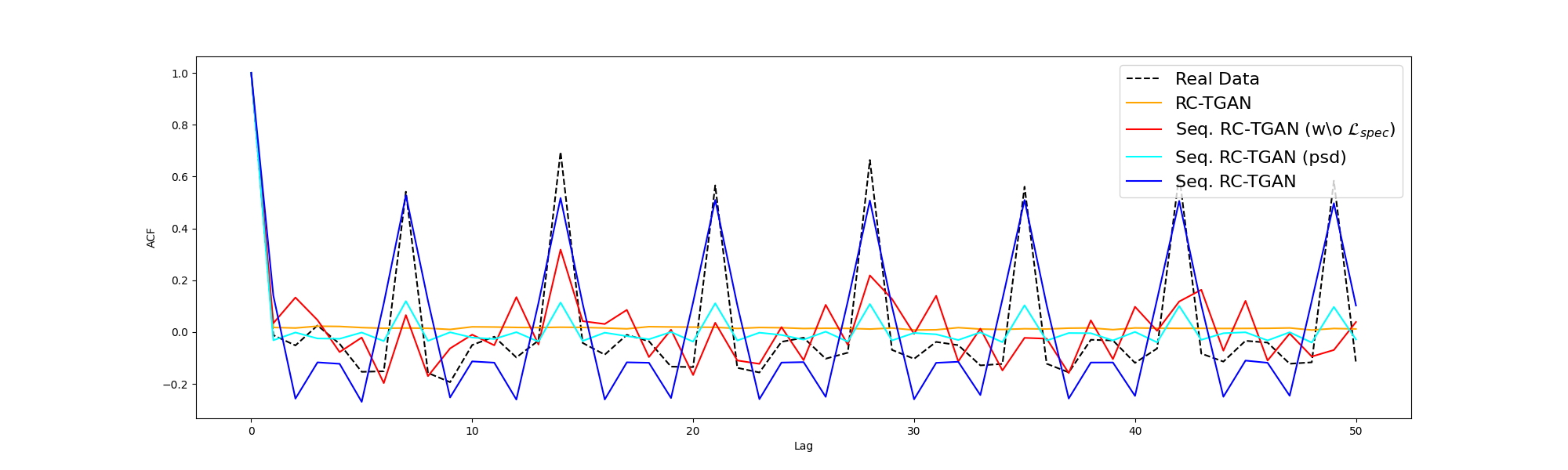}
        \caption{Rossmann Store Sales: Ablation Study}
        \label{fig:acf_rossman_ablation}
    \end{subfigure}
    \hfill
    \begin{subfigure}[b]{0.47\textwidth}
        \centering
        \includegraphics[width=\textwidth]{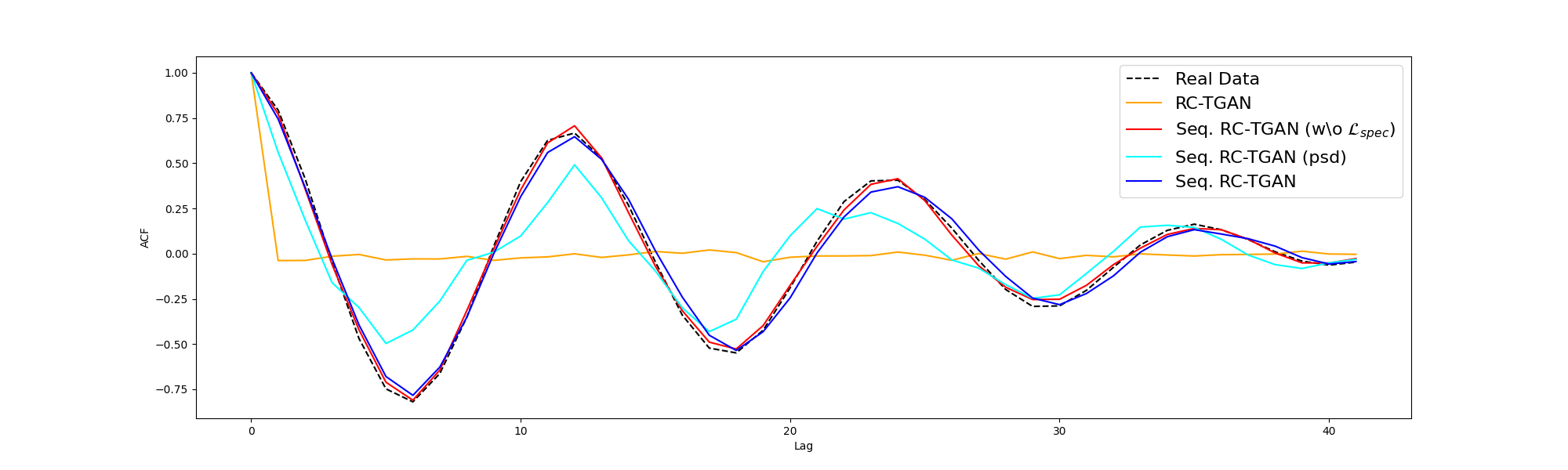}
        \caption{Walmart Fuel Price: Ablation Study}
        \label{fig:acf_walmart_ablation}
    \end{subfigure}
    
    \caption{\textbf{Autocorrelation Function (ACF) Analysis.} The dashed black line represents the ground truth. \textbf{Top Row (a, b):} Comparison against baselines. Static models like SDV (pink) and ClavaDDPM (green) fail to capture seasonality (recurring peaks at lag 7 for Rossman and lag 12 for Walmart). \textbf{Bottom Row (c, d):} Ablation study. The static RC-TGAN (orange) produces a flat line, and the unregularized \textit{Seq. RC-TGAN (no $\mathcal{L}_{spec}$)} (red) captures local transitions but underestimates global amplitudes. The proposed \textit{Seq. RC-TGAN} (blue) accurately reproduces the overarching seasonal correlation structure.}
    \label{fig:acf_grid_analysis}
\end{figure*}

This dynamic is visually corroborated by the Autocorrelation Function (ACF) analysis of numerical time series presented in Figure \ref{fig:acf_grid_analysis}. As seen in the top row (a, b), static tabular models like SDV and ClavaDDPM completely fail to capture temporal dependencies, resulting in flat ACF curves. This visual analysis highlights a critical limitation of the commonly used MSE (ACF) metric. On the Rossmann dataset, the static SDV model achieves a nominally better MSE (ACF) compared to the sequential DoppelGANger model (0.0700 vs. 0.0868). However, visual inspection reveals that SDV merely produces a flat line near zero; this mathematically minimizes the mean squared error across all lags by "playing it safe," but fails entirely to capture underlying temporal dynamics. DoppelGANger successfully reproduces the oscillating shape of the sales autocorrelation but misses the exact amplitude of certain peaks, resulting in a harsher point-wise MSE penalty. 

The spectral metrics ($\overline{\mathcal{D}}_{spec}$ and $\overline{\mathcal{D}}_{env}$) correct this incompleteness by evaluating the global frequency structure rather than local point-wise errors. Ultimately, our proposed Seq. RC-TGAN framework bridges all these gaps, demonstrating the ability to accurately capture local point-wise dependencies, global continuous periodicities via VGM adaptation, and crucially, the complex structural harmonics inherent to categorical time series.

\begin{table}[htbp]
    \centering
    \caption{Perfomance on simulated datasets: Comparison of $\overline{\mathcal{D}}_{env}$ between our proposed method and baselines across different state space sizes ($K$). Lower is better.}
    \label{tab:metrics_simulated_data}
    \resizebox{\columnwidth}{!}{%
    \begin{tabular}{lccc}
        \toprule
        & \multicolumn{3}{c}{\textbf{State Space Size ($K$)}} \\
        \cmidrule(lr){2-4}
        \textbf{Model} & \textbf{K=7} & \textbf{K=12} & \textbf{K=21} \\
        \midrule
        \multicolumn{4}{l}{\textbf{\textit{Noisy Cyclic}}} \\
        \midrule
        SDV                    & $\underline{0.7830 \pm 0.0582}$ & $0.9459 \pm 0.1804$ & $0.8418 \pm 0.0794$ \\
        ClavaDDPM              & $0.8729 \pm 0.0660$ & $0.8922 \pm 0.1036$ & $0.9101 \pm 0.1797$ \\
        DoppelGANger           & $0.9467 \pm 0.0579$ & $\underline{0.8786 \pm 0.1071}$ & $\underline{0.8336 \pm 0.0452}$ \\
        TimeGAN                & $0.9100 \pm 0.1302$ & $0.9135 \pm 0.1427$ & $0.9076 \pm 0.0860$ \\
        Seq. RC-TGAN           & $\mathbf{0.4246 \pm 0.0117}$ & $\mathbf{0.4984 \pm 0.0319}$ & $\mathbf{0.5988 \pm 0.0202}$ \\
        \midrule
        \multicolumn{4}{l}{\textbf{\textit{Symmetric Sticky}}} \\
        \midrule
        SDV                    & $0.2182 \pm 0.0297$ & $0.1012 \pm 0.0144$ & $0.0661 \pm 0.0112$ \\
        ClavaDDPM              & $0.1873 \pm 0.0266$ & $\underline{0.0949 \pm 0.0050}$ & $0.0569 \pm 0.0065$ \\
        DoppelGANger           & $0.7691 \pm 0.1404$ & $0.2123 \pm 0.0240$ & $0.1258 \pm 0.0173$ \\
        TimeGAN                & $\underline{0.1813 \pm 0.0208}$ & $0.1007 \pm 0.0069$ & $\underline{0.0555 \pm 0.0074}$ \\
        Seq. RC-TGAN           & $\mathbf{0.1221 \pm 0.0023}$ & $\mathbf{0.0448 \pm 0.0018}$ & $\mathbf{0.0347 \pm 0.0009}$ \\
        \bottomrule
    \end{tabular}
    }
\end{table}

\begin{table}[htbp]
    \centering
    \caption{Performance Metrics on real-world datasets: MSE (ACF), $\overline{\mathcal{D}}_{spec}$, and $\overline{\mathcal{D}}_{env}$}
    \small
    \resizebox{\columnwidth}{!}{%
    \begin{tabular}{llccc}
        \toprule
        \textbf{Dataset} & \textbf{Model} & \textbf{MSE (ACF)} & \textbf{$\overline{\mathcal{D}}_{spec}$} & \textbf{$\overline{\mathcal{D}}_{env}$} \\
        \midrule
        \multirow{5}{*}{\textbf{Rossmann}} 
        & SDV                    & $\underline{0.0700 \pm 0.0000}$ & $50.00\% \pm 0.00\%$ & $\underline{0.7359 \pm 0.0000}$ \\
        & ClavaDDPM              & $0.0702 \pm 0.0001$ & $50.30\% \pm 0.08\%$ & $0.7407 \pm 0.0006$ \\
        & DoppelGANger           & $0.0868 \pm 0.0293$ & $\underline{46.22\% \pm 4.34\%}$ & $0.7792 \pm 0.5640$ \\
        & TimeGAN                & $0.0951 \pm 0.0317$ & $57.20\% \pm 5.21\%$ & $1.2507 \pm 0.3600$ \\
        & Seq. RC-TGAN & $\mathbf{0.0340 \pm 0.0072}$ & $\mathbf{30.66\% \pm 3.86\%}$ & $\mathbf{0.4578 \pm 0.1630}$ \\
        \midrule
        \multirow{5}{*}{\textbf{Walmart}} 
        & SDV                    & $0.1223 \pm 0.0000$ & $50.99\% \pm 0.00\%$ & $0.0757 \pm 0.0000$ \\
        & ClavaDDPM              & $0.1195 \pm 0.0015$ & $45.04\% \pm 0.40\%$ & $0.0727 \pm 0.0008$ \\
        & DoppelGANger           & $\mathbf{0.0120 \pm 0.0052}$ & $\underline{6.88\% \pm 0.89\%}$ & $\underline{0.0118 \pm 0.0028}$ \\
        & TimeGAN                & $0.1250 \pm 0.0208$ & $16.32\% \pm 2.90\%$ & $0.1446 \pm 0.0108$ \\
        & Seq. RC-TGAN & $\underline{0.0130 \pm 0.0112}$ & $\mathbf{5.17\% \pm 0.58\%}$ & $\mathbf{0.0048 \pm 0.0023}$ \\
        \bottomrule
    \end{tabular}
    }
    \label{tab:metrics_real_data}
\end{table}

\subsection{Ablation Study}

To rigorously isolate the contributions of our architectural design choices, specifically the recurrent temporal generation and the proposed frequency-domain loss, we conducted an ablation study. We compared four variants of our framework: RC-TGAN \cite{gueye_row_2023} (no temporal dimension modeling), the Seq. RC-TGAN (w\textbackslash o $\mathcal{L}_{spec}$)  that is the recurrent baseline without any spectral loss, the Seq. RC-TGAN (psd) which models based on the spectral density loss for the numerical columns instead of the spectral envelope loss (based on VGM), and our proposed full model, Seq. RC-TGAN.

The necessity of the defined spectral loss is most starkly evident in our highly controlled simulated environments (Table \ref{tab:metrics_simulated_data_ablation}). While the model Seq. RC-TGAN (w\textbackslash o $\mathcal{L}_{spec}$) demonstrates improvements on empirical real-world data, it completely fails to reproduce pure mathematical periodicities. The results show that without explicit frequency-domain guidance, the recurrent baseline performs almost identically to the static RC-TGAN model on both the NCP and SSP across all state space sizes. 

This reveals that standard adversarial training in the time domain, even with an RNN-based architecture, is insufficient to prevent white-noise-like spectra when faced with  periodic constraints. By incorporating the spectral envelope loss, the divergence ($\overline{\mathcal{D}}_{env}$) is reduced by approximately 50\% across almost all configurations (e.g., a statistically significant divergence drop from 0.8483 to 0.4246 for the NCP at $K=7$). These results confirm that the proposed spectral loss is not merely an incremental tuning parameter for real-world data, but a fundamentally essential component for generative models to successfully reconstruct latent harmonics and system inertia in categorical time series.

Table \ref{tab:metrics_real_data_ablation} details the performance of these variants on the Rossmann and Walmart datasets. Transitioning from a static generator (RC-TGAN) to a recurrent architecture (Seq. RC-TGAN (w/o $\mathcal{L}_{spec}$)) yields statistically significant improvements ($p < 0.05$) across both time and frequency domains on real-world data. For instance, on the Walmart dataset, introducing the recurrent structure reduces the time-domain MSE (ACF) by 54.6\% (from 0.1200 to 0.0545) and drastically reduces the continuous frequency divergence ($\overline{\mathcal{D}}_{spec}$) by 81.8\% (from 45.26\% to 8.22\%). 

Evaluating the progression of our ablation study highlights the fundamental necessity of the spectral envelope loss for mixed-type columns in  relational databases. First, comparing the unregularized recurrent baseline (Seq. RC-TGAN (w\textbackslash o $\mathcal{L}_{spec}$)) to the Seq. RC-TGAN (psd) variant demonstrates the advantage of the latter, as applying a standard spectral density loss successfully provides frequency-domain guidance for continuous numerical columns. However, this approach remains fundamentally insufficient for reliably capturing the complex periodic dynamics inherent to categorical data. Subsequently, comparing the Seq. RC-TGAN (psd) variant to our full model (Seq. RC-TGAN) illustrates the critical impact of our proposed approach. Implementing the full Spectral Envelope loss forces the generator to comprehensively learn overarching periodic patterns across all data types natively. This yields a statistically significant 23.9\% relative reduction in categorical SED ($\overline{\mathcal{D}}_{env}$) compared to the psd variant on the Rossmann dataset (dropping from 0.6014 to 0.4578). Furthermore, on the Walmart dataset, the full model drives the SED down to 0.0048, outperforming the PSD-only variant and achieving a massive 93.4\% overall improvement compared to the initial recurrent baseline (dropping from 0.0732 to 0.0048).

These findings are visually confirmed by the Autocorrelation Function (ACF) analysis (Figure \ref{fig:acf_grid_analysis}, bottom row). The static RC-TGAN completely fails to capture temporal dependencies, resulting in a flat ACF line. While the Seq. RC-TGAN (w\textbackslash o $\mathcal{L}_{spec}$) successfully begins to capture localized temporal transitions, it still misses the global structural amplitudes. The integration of the full spectral loss acts as the definitive catalyst, enabling the generator to accurately reproduce long-range seasonal correlations rather than just step-by-step localized transitions.

\begin{table}[htbp]
    \centering
    \caption{Ablation study on simulated datasets: Comparison of $\overline{\mathcal{D}}_{env}$ between RC-TGAN variants across different state space sizes ($K$). Lower is better.}
    \label{tab:metrics_simulated_data_ablation}
    \resizebox{\columnwidth}{!}{%
    \begin{tabular}{lccc}
        \toprule
        & \multicolumn{3}{c}{\textbf{State Space Size ($K$)}} \\
        \cmidrule(lr){2-4}
        \textbf{Model} & \textbf{K=7} & \textbf{K=12} & \textbf{K=21} \\
        \midrule
        \multicolumn{4}{l}{\textbf{\textit{Benchmark: Noisy Cyclic}}} \\
        \midrule
        RC-TGAN      & $\underline{0.8215 \pm 0.0864}$ & $\underline{0.9719 \pm 0.1127}$ & $0.9652 \pm 0.3088$ \\
        Seq. RC-TGAN (w\textbackslash o $\mathcal{L}_{spec}$)         & $0.8483 \pm 0.1094$ & $0.9775 \pm 0.1155$ & $\underline{0.8260 \pm 0.0261}$ \\
        Seq. RC-TGAN & $\mathbf{0.4246 \pm 0.0117}$ & $\mathbf{0.4984 \pm 0.0319}$ & $\mathbf{0.5988 \pm 0.0202}$ \\
        \midrule
        \multicolumn{4}{l}{\textbf{\textit{Benchmark: Symmetric Sticky}}} \\
        \midrule
        RC-TGAN      & $\underline{0.1962 \pm 0.0159}$ & $0.0957 \pm 0.0049$ & $\underline{0.0546 \pm 0.0065}$ \\
        Seq. RC-TGAN (w\textbackslash o $\mathcal{L}_{spec}$)             & $0.1988 \pm 0.0169$ & $\underline{0.0926 \pm 0.0086}$ & $0.0588 \pm 0.0029$ \\
        Seq. RC-TGAN & $\mathbf{0.1221 \pm 0.0023}$ & $\mathbf{0.0448 \pm 0.0018}$ & $\mathbf{0.0347 \pm 0.0009}$ \\
        \bottomrule
    \end{tabular}
    }
\end{table}

\begin{table}[htbp]
    \centering
    \caption{Ablation study on real-world datasets: MSE (ACF), $\overline{\mathcal{D}}_{spec}$, and $\overline{\mathcal{D}}_{env}$}
    \small
    \resizebox{\columnwidth}{!}{%
    \begin{tabular}{llccc}
        \toprule
        \textbf{Dataset} & \textbf{Model} & \textbf{MSE (ACF)} & \textbf{$\overline{\mathcal{D}}_{spec}$} & \textbf{$\overline{\mathcal{D}}_{env}$} \\
        \midrule
        \multirow{3}{*}{\textbf{Rossmann}} 
        & RC-TGAN      & $0.0703 \pm 0.0002$ & $50.29\% \pm 0.06\%$ & $0.7404 \pm 0.0003$ \\
        & Seq. RC-TGAN (w\textbackslash o $\mathcal{L}_{spec}$)           & $0.0687 \pm 0.0037$ & $\underline{43.72\% \pm 2.54\%}$ & $\underline{0.5790 \pm 0.1574}$ \\
        & Seq. RC-TGAN (psd)           & $\underline{0.0681 \pm 0.0322}$ & $43.99\% \pm 4.78\%$ & $0.6014 \pm 0.1771$ \\
        & Seq. RC-TGAN & $\mathbf{0.0340 \pm 0.0072}$ & $\mathbf{30.66\% \pm 3.86\%}$ & $\mathbf{0.4578 \pm 0.1630}$ \\
        \midrule
        \multirow{3}{*}{\textbf{Walmart}} 
        & RC-TGAN      & $0.1200 \pm 0.0012$ & $45.26\% \pm 0.58\%$ & $0.0833 \pm 0.0017$ \\
        & Seq. RC-TGAN (w\textbackslash o $\mathcal{L}_{spec}$)           & $0.0545 \pm 0.0227$ & $\underline{8.22\% \pm 0.96\%}$ & $0.0732 \pm 0.0016$ \\
        & Seq. RC-TGAN (psd)           & $ \underline{0.0389\pm 0.0212}$ & $16.23\% \pm 8.16\%$ & $ \underline{0.0067\pm 0.0102}$ \\
        & Seq. RC-TGAN & $\mathbf{0.0130 \pm 0.0112}$ & $\mathbf{5.17\% \pm 0.58\%}$ & $\mathbf{0.0048 \pm 0.0023}$ \\
        \bottomrule
    \end{tabular}
    }
    \label{tab:metrics_real_data_ablation}
\end{table}

\section{Conclusion} \label{sec:conclusion}
In this paper, we addressed the critical challenge of generating high-fidelity time series within relational databases by introducing Seq. RC-TGAN, a sequential generative adversarial network enhanced with a novel, integrated spectral envelope loss. Rather than relying solely on static encodings, our framework explicitly optimizes the network to preserve the complex frequency-domain features of both categorical and continuous time series during training. Furthermore, we established a mathematically rigorous evaluation paradigm by analytically deriving the spectral envelope for circulant Markov chains, providing a "gold standard" for categorical time series alongside two novel spectral divergence metrics. Extensive experiments on these simulated data and real-world datasets (Rossmann and Walmart) demonstrate that our approach significantly outperforms state-of-the-art baselines in capturing latent periodicities, strict cyclic constraints, and long-term seasonality.

\bibliographystyle{IEEEtran}
\bibliography{references}

\appendix
\section*{Proof of Lemma \ref{lemma:specenv_continuity}}
By assumption, the spectral density matrix of the one-hot encoded process, $f_Y(\omega)$, is continuous. 
Because the categories are mutually exclusive and exhaustive, the covariance matrix $V$ has rank $K-1$. We can apply any $K \times (K-1)\text{-dim}$ projection matrix $Q$ to obtain a full-rank covariance matrix $\overline{V} = Q^{\prime} V Q$ and a projected spectral density matrix $\overline{f}_Y(\omega) = Q^{\prime} f_Y(\omega) Q$. 

The spectral envelope $\lambda(\omega)$ is given by the largest eigenvalue of the matrix $C(\omega) = \overline{V}^{-1/2} \overline{f}_Y(\omega) \overline{V}^{-1/2}$. 
Since $f_Y(\omega)$ is continuous with respect to $\omega$, the matrix $C(\omega)$ is also continuous. 

The eigenvalues of $C(\omega)$ are the roots of its characteristic polynomial, which can be defined as $P(x, \omega) = \det(x I - C(\omega))$. The coefficients of this polynomial are continuous functions of the entries of $C(\omega)$ which are continuous functions of $\omega$. Then, polynomial $P(x, \omega)$ is continuous function of $\omega$. 

According to standard mathematical theorem in \cite{ross2022yet} regarding the roots of polynomials, the roots of a polynomial are continuous functions of its coefficients. Consequently, the largest root, $\lambda(\omega)$, is a continuous function on the fundamental frequency domain $[-1/2, 1/2]$.\\

\section*{Proof of Lemma \ref{lemma:specenv_boundness}}
We prove the two properties sequentially based on the definitions of the spectral envelope and the norms.

Let $\overline{f}_Y(\omega)$ be the projected spectral density matrix of dimension $(K-1) \times (K-1)$, and let $\overline{V}$ be the corresponding full-rank covariance matrix. The integral of the spectral density matrix over the fundamental frequency domain yields the covariance matrix:
$$
    \int_{-1/2}^{1/2} \overline{f}_Y(\omega) d\omega = \overline{V}.
$$
Recall that the spectral envelope is defined as $\lambda(\omega) = \mu_{\max}(C(\omega))$, where $\mu_{\max}(\cdot)$ is the highest eigenvalue function and $C(\omega) = \overline{V}^{-1/2} \overline{f}_Y(\omega) \overline{V}^{-1/2}$. 

\textit{Upper Bound:} Integrating $C(\omega)$ over the frequency domain yields:
\begin{align*}
\int_{-1/2}^{1/2} C(\omega) d\omega & = \overline{V}^{-1/2} \left( \int_{-1/2}^{1/2} \overline{f}_Y(\omega) d\omega \right) \overline{V}^{-1/2} \\
& = I_{K-1}. 
\end{align*}
Taking the trace of both sides, we get:
\begin{align*}
\int_{-1/2}^{1/2} \text{tr}(C(\omega)) d\omega & = \text{tr}\left(\int_{-1/2}^{1/2} C(\omega) d\omega\right) \\
& = \text{tr}(I_{K-1}) = K-1.
\end{align*}
Since $C(\omega)$ is positive semi-definite, its maximum eigenvalue is bounded by its trace, $\mu_{\max}(C(\omega)) \leq \text{tr}(C(\omega))$ for all $\omega$. Therefore:
$$
    \Vert \lambda \Vert_1 = \int_{-1/2}^{1/2} \lambda(\omega) d\omega \leq \int_{-1/2}^{1/2} \text{tr}(C(\omega)) d\omega = K-1.
$$

\textit{Lower Bound:} Using the variational characterization of the spectral envelope, for any non-zero vector $\bar{\beta}_0 \in \mathbb{R}^{K-1}$, we have:
$$
    \lambda(\omega) = \sup_{\bar{\beta} \in \mathbb{R}^{K-1}} \frac{\bar{\beta}^{\prime} \overline{f}_Y(\omega) \bar{\beta}}{\bar{\beta}^{\prime} \overline{V} \bar{\beta}} \geq \frac{\bar{\beta}_0^{\prime} \overline{f}_Y(\omega) \bar{\beta}_0}{\bar{\beta}_0^{\prime} \overline{V} \bar{\beta}_0}.
$$
Integrating this inequality over the frequency domain gives:

\begin{align*}
    \Vert \lambda \Vert_1 & = \int_{-1/2}^{1/2} \lambda(\omega) d\omega \\
    & \geq \int_{-1/2}^{1/2} \frac{\bar{\beta}_0^{\prime} \overline{f}_Y(\omega) \bar{\beta}_0}{\bar{\beta}_0^{\prime} \overline{V} \bar{\beta}_0} d\omega \\
    & = \frac{\bar{\beta}_0^{\prime} \left( \int_{-1/2}^{1/2} \overline{f}_Y(\omega) d\omega \right) \bar{\beta}_0}{\bar{\beta}_0^{\prime} \overline{V} \bar{\beta}_0} \\
    & = \frac{\bar{\beta}_0^{\prime} \overline{V} \bar{\beta}_0}{\bar{\beta}_0^{\prime} \overline{V} \bar{\beta}_0} = 1.
\end{align*}
Thus, $1 \leq \Vert \lambda \Vert_1 \leq K-1$.

From Lemma \ref{lemma:specenv_continuity}, $\lambda(\omega)$ is a continuous function on the compact interval $[-1/2, 1/2]$. Therefore, it is bounded, which implies $\lambda \in L^\infty([-1/2, 1/2])$ and consequently $\lambda \in L^2([-1/2, 1/2])$, meaning $\Vert \lambda \Vert_2 < \infty$.

For the lower bound, we apply Jensen's inequality (or the Cauchy-Schwarz inequality) on the probability space defined by the interval $[-1/2, 1/2]$ with length 1:
$$
    \Vert \lambda \Vert_2^2 = \int_{-1/2}^{1/2} \lambda(\omega)^2 d\omega \geq \left( \int_{-1/2}^{1/2} \lambda(\omega) d\omega \right)^2 = \Vert \lambda \Vert_1^2.
$$
Since we established in Part (i) that $\Vert \lambda \Vert_1 \geq 1$, it strictly follows that $\Vert \lambda \Vert_2 \geq 1$. \\

\section*{Proof of Lemma \ref{lemma:circulant_envelope}}

Because $P$ is a circulant matrix, it is a normal matrix ($PP^{\prime} = P^{\prime} P$). A fundamental property of circulant matrices is that they are diagonalized by the Discrete Fourier Transform (DFT) matrix. Therefore, the eigenvectors $v_k$ are the fixed, orthogonal Fourier basis vectors. This orthogonality allows us to project the multivariate one-hot encoded categorical process $Y_t$ into $K$ uncorrelated scalar processes, defined as $Z_t^{(k)} = v_k^* Y_t$.\\

The conditional expectation of each projected scalar process is exactly governed by its corresponding eigenvalue: $\mathbb{E}[Z_{t+1}^{(k)} \mid Z_t^{(k)}] = \gamma_k Z_t^{(k)}$. This equation defines a complex Autoregressive Process of order 1 (AR(1)). For such an AR(1) process, the temporal dependence decays geometrically. Consequently, the normalized autocorrelation function at lag $h$ is given exactly by the corresponding eigenvalue raised to the absolute lag: $R_k(h) = \gamma_k^{|h|}$.\\

By the Wiener-Khinchin theorem, the spectral density $f_k(\omega)$ of a stationary discrete-time process is the Discrete-Time Fourier Transform (DTFT) of its autocorrelation sequence. Substituting the geometric autocorrelation $R_k(h) = \gamma_k^{|h|}$ into the Fourier sum yields an explicitly solvable infinite geometric series:

\begin{align*}
f_k(\omega) & = \sum_{h=-\infty}^{\infty} R_k(h) e^{-2\pi i\omega h} \\
& = \sum_{h=-\infty}^{\infty} \gamma_k^{|h|} e^{-2\pi i\omega h} = \frac{1 - |\gamma_k|^2}{|1 - \gamma_k e^{-2\pi i\omega}|^2}.
\end{align*}

Using the polar representation of the eigenvalue $\gamma_k = r_k e^{i\phi_k}$, we can expand the squared norm in the denominator:
$$
    |1 - r_k e^{i(\phi_k - 2\pi\omega)}|^2 = 1 - 2r_k \cos(2\pi\omega - \phi_k) + r_k^2
$$
This yields the explicit polar form for the spectral density: $f_k(\omega) = \frac{1 - r_k^2}{1 - 2r_k \cos(2\pi\omega - \phi_k) + r_k^2}$.\\

The spectral envelope is defined as the supremum of the normalized spectral density over all possible projection vectors $\beta$: $\lambda(\omega) = \sup_{\beta} \frac{f_Z(\omega; \beta)}{\text{Var}(Z)}$. Because the scalar modes $Z_t^{(k)}$ are mutually uncorrelated (their cross-covariance is zero for all lags), the total spectral density of any linear combination is simply the sum of the individual mode densities. 

To maximize a weighted average of independent components at any specific frequency $\omega$, the optimal strategy is to assign all weight to the single component with the largest value. Thus, evaluating the envelope simplifies to finding the point-wise maximum of the individual harmonic densities (excluding the trivial stationary DC component $k=0$):
$$
    \lambda(\omega) = \max_{k \in \{1, \dots, K-1\}} f_k(\omega)
$$
\end{document}